\documentclass{article}
\usepackage[T1]{fontenc}   
\usepackage{xcolor}   

\usepackage[preprint]{corl_2026}   

\usepackage{booktabs}
\usepackage{multirow}
\usepackage{array}
\usepackage{graphicx}
\usepackage{amsmath,amssymb}
\usepackage{xspace}
\usepackage{xcolor}
\usepackage{tikz}
\usetikzlibrary{arrows.meta,calc,positioning,shapes.geometric}
\usepackage{adjustbox}
\usepackage{enumitem}
\setlist{nosep,leftmargin=*}

\newcommand{\rig}{LIBERO-RIG\xspace}

\newcommand{\pizero}{$\pi_0$\xspace}
\newcommand{\pihalf}{$\pi_{0.5}$\xspace}

\definecolor{axobs}{HTML}{5C6F7E}\definecolor{axlat}{HTML}{6F8A6F}\definecolor{axweak}{HTML}{B08A7A}
\newcommand{\axncolor}[1]{\ifnum#1<5 axobs\else\ifnum#1<9 axlat\else\ifnum#1<11 axobs\else axweak\fi\fi\fi}
\DeclareRobustCommand{\axn}[1]{\tikz[baseline=(c.base)]{\node[draw=\axncolor{#1},text=\axncolor{#1},circle,inner sep=0.15pt,minimum size=1.12em,text height=1.3ex,text depth=0pt,line width=0.45pt,font=\scriptsize] (c) {#1};}}
\DeclareRobustCommand{\axnt}[1]{\tikz[baseline={([yshift=-0.3ex]c.base)}]{\node[draw=\axncolor{#1},text=\axncolor{#1},circle,inner sep=0.1pt,minimum size=0.9em,text height=1.0ex,text depth=0pt,line width=0.4pt,font=\fontsize{6}{6}\selectfont] (c) {#1};}}

\newcommand{\appref}[1]{Appendix~\ref{#1}}

\newcommand{\rot}[1]{\rotatebox{60}{\footnotesize #1}}

\newcounter{finding}
\newcommand{\finding}[1]{\refstepcounter{finding}\vspace{2pt}\noindent\textbf{Finding \thefinding: #1}}

\newcommand{\pooledNominal}{0.932}
\newcommand{\pooledSa}{0.914}
\newcommand{\pooledSb}{0.845}
\newcommand{\pooledSc}{0.767}
\newcommand{\pooledSd}{0.591}
\newcommand{\pooledSe}{0.509}

\newcommand{\axPtwo}{0.448}

\newcommand{\axAfive}{0.931}

\newcommand{\ctrlFiveHz}{0.153}

\newcommand{\episodesTotal}{881,500}
\newcommand{\episodesPerArm}{107,000}

\newcommand{\ratioa}{0.981}
\newcommand{\ratioLoa}{0.978}
\newcommand{\ratioHia}{0.984}

\newcommand{\ratiob}{0.907}
\newcommand{\ratioLob}{0.896}
\newcommand{\ratioHib}{0.916}
\newcommand{\ciLoc}{0.742}
\newcommand{\ciHic}{0.789}

\newcommand{\sigPairs}{35}
\newcommand{\letterGroups}{7}
\newcommand{\letterGroupsSfour}{7}
\newcommand{\letterGroupsSfive}{5}

\newcommand{\trendBlocks}{304}
\newcommand{\trendNeg}{287}
\newcommand{\trendConst}{12}
\newcommand{\trendNonInc}{222}
\newcommand{\trendStrict}{193}
\newcommand{\trendExceptions}{17}
\newcommand{\trendMeanRho}{-0.905}

\newcommand{\rqfourP}{0.008}

\usepackage[capitalize]{cleveref}  
\usepackage{flafter}               
\usepackage{float}                 

\title{\rig: A Hardware-Setup Robustness Benchmark for Vision-Language-Action Models}

\author{\mdseries   
  Andrea Yaoyun Cui\quad Hanyu Wang\quad Xinyi Wang\quad Bangzheng Li\quad Lin Zhang\\
  Zhenggang Tang\quad Shyamal Buch\quad Dejia Xu\quad Matthew Tancik\\
  Luma AI
}

\begin{document}
\maketitle
\begin{abstract}
Robot policies are expected to generalize across deployment environments, including variations in the robot itself: cameras are bumped and remounted, calibrations drift, controllers run at different rates, and end-effector pads wear down. We present \rig, a benchmark that perturbs only the robot’s hardware and setup while holding every LIBERO task fixed. The benchmark spans eleven perturbation axes, including camera mounting and optics, camera-to-arm calibration, actuation gain, control rate, proprioceptive noise, base pose, table height, and gripper pads. Each axis has five severity levels, yielding 210 conditions across the four LIBERO suites. We evaluate nine recent vision-language-action models using 500 episodes per condition and substantiate each finding with rigorous statistical tests. Pooled success falls from \pooledNominal{} to \pooledSc{} at the middle severity and \pooledSe{} at the highest severity. Performance degradation varies substantially across perturbation axes and action interfaces, and training on visual perturbations does not mitigate faults that leave camera images unchanged. We release the benchmark together with manifests specifying the fixed perturbation parameters for each episode.

\end{abstract}


\begin{figure}[H]
\centering
\includegraphics[width=0.86\linewidth]{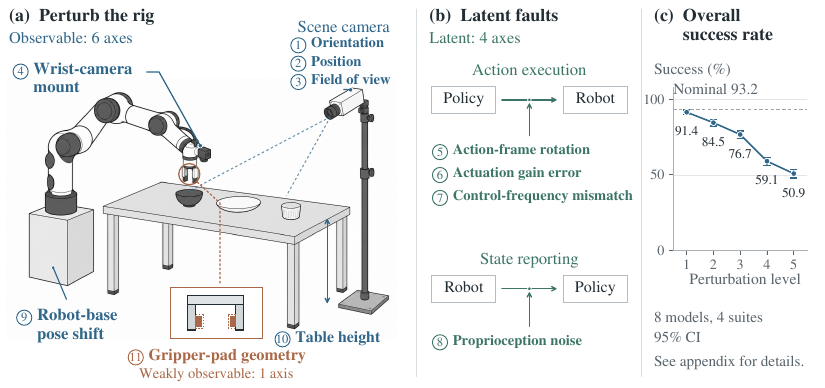}
\caption{\textbf{Same task, perturbed rig.} The instruction, object identities, goal, initial state, and success predicate remain fixed while one of 11 rig properties varies across five severity levels. (a) Six observable axes (blue) alter camera views of the same arm and object configuration; changes in gripper-pad geometry (orange) are visible only in close-range wrist-camera views. (b) Four latent axes (green) alter command execution or reported state without changing the image. (c) Overall success at each severity level for the eight models evaluated on all four suites, with 95\% intervals. Per-suite results appear in \cref{fig:rq2}, numerical values in \cref{tab:severity}, and severity definitions in \cref{tab:axes}. Circled numbers \axnt{1}--\axnt{11} identify the axes consistently across all tables and figures.}
\label{fig:teaser}
\end{figure}

\section{Introduction}
\label{sec:intro}

A learned robot policy is expected to generalize beyond its training conditions, including variations in the robot itself. The robot executing a policy often differs from the one used to collect its demonstrations: cameras are bumped and remounted, camera-to-arm calibration drifts, control rates change, gripper pads wear, and the arm may be mounted on a different table. The task remains the same; what changes is the rig, the hardware beneath the policy and the way it is set up.

Vision-language-action (VLA) models~\cite{brohan2023rt2,kim2024openvla,black2024pi0} are the dominant approach to such policies, and LIBERO~\cite{liu2023libero} is a standard benchmark in which the best open models achieve success rates of 93-98\% of episodes. Existing robustness benchmarks built on LIBERO and other simulators perturb the task~\cite{zhou2025liberopro}, visual observations~\cite{fei2026liberoplus}, or scene factors such as color, lighting, and distractors~\cite{pumacay2024colosseum,xie2023factorworld}. Their rig perturbations are limited to the camera viewpoint, the arm's initial configuration, and the table position; none evaluates camera-to-arm calibration, actuation gain, control rate, the wrist camera, proprioception, or the end effector. A policy can therefore appear robust across existing benchmarks while remaining dependent on a specific rig configuration.

We propose \rig, which perturbs the rig along 11 axes, each at five severity levels ranging from everyday drift to a markedly different setup, yielding 210 conditions across the four LIBERO suites (\cref{fig:teaser}). We evaluate nine VLA models spanning four action-decoding families using 500 episodes per condition, for a total of \episodesTotal{} episodes. To our knowledge, \rig is the first robustness benchmark for VLA models that targets the rig itself; it gives each model a per-axis robustness profile rather than a single score, revealing failure patterns associated with model design.

\section{The \rig Benchmark}
\label{sec:benchmark}
\begin{table*}[b]
\centering
\caption{\textbf{The 11 perturbation axes.} Circled numbers match those in \cref{fig:teaser}. Fault types are defined by the fault's direct effect on the camera images at a fixed arm and object configuration: O observable, L latent, W weakly observable (visible only in close-range wrist views). Each severity ladder runs from s1 to s5; \appref{sec:supp_axes} lists the parameters at each level.}
\label{tab:axes}
\small
\begin{adjustbox}{max width=\textwidth}\begin{tabular}{@{}r l c p{9.6cm} l@{}}
\toprule
\# & Axis & Type & What physically changes & Ladder s1 to s5 \\
\midrule
\axn{1} & camera orientation & O & the fixed scene camera's mount is twisted (pitch, yaw, roll together); its position does not move & 2 to 25 degrees \\
\axn{2} & camera position & O & the scene camera is remounted elsewhere on a sphere around the workspace, still aimed at it & 5 to 50 degrees of arc, dolly up to 1.6$\times$ \\
\axn{3} & camera field of view & O & a different lens: the vertical field of view changes from the nominal 45 degrees & $\pm$2 to $\pm$18 degrees \\
\axn{4} & wrist-camera mount & O & the wrist camera's bracket is twisted and slid along its mounting arm & 1 to 15 degrees, 2 to 30 mm \\
\axn{5} & action-frame rotation & L & camera-to-arm (hand-eye) miscalibration: every command executes in a frame rotated about vertical & 2 to 25 degrees \\
\axn{6} & actuation gain error & L & commands execute too strong or too weak by a scale factor & $\pm$5 to $\pm$50 percent \\
\axn{7} & control-frequency mismatch & L & the control loop runs at a different rate than the trained 20 Hz & 25, 10, 50, 5, 100 Hz \\
\axn{8} & proprioception noise & L & Gaussian noise on the sensed hand pose reported to the policy; the controller still reads the true state & 2 to 50 mm, 0.5 to 8 degrees \\
\axn{9} & robot-base pose shift & O & the floor mount is moved and rotated; the objects stay where they are & 1 cm and 2 degrees to 15 cm and 25 degrees \\
\axn{10} & table-height change & O & the work surface is raised or lowered; objects move with it (two suites) & $\pm$1 to $\pm$10 cm \\
\axn{11} & gripper-pad geometry & W & the end-effector pads are worn down or oversized; fingers and joints are untouched & $\pm$5 to $\pm$25 percent pad scale \\
\bottomrule
\end{tabular}\end{adjustbox}
\end{table*}

\textbf{Problem and rig.} The rig is parameterized by the scene camera's \axnt{1}~orientation, \axnt{2}~position, and \axnt{3}~field of view, \axnt{4}~the wrist camera's mounting pose, \axnt{5}~the camera-to-arm calibration, the controller's \axnt{6}~gain and \axnt{7}~rate, \axnt{8}~the noise in the arm's reported state, \axnt{9}~the arm's base pose, \axnt{10}~the table height, and \axnt{11}~the end-effector pad geometry. LIBERO~\cite{liu2023libero} simulates a Franka Panda arm in MuJoCo~\cite{todorov2012mujoco} using robosuite~\cite{zhu2020robosuite}; \rig keeps each task's instruction, object identities, success predicate, and nominal initial states fixed while changing only the rig.

\textbf{Axes.} \cref{tab:axes} lists the 11 axes, their corresponding physical changes, and the severity endpoints. We classify each fault by its direct effect on camera images at a fixed arm and object configuration: an observable fault changes at least one image, whereas a latent fault leaves all images unchanged and affects the policy only through the arm's motion or its reported state.

\textbf{Conditions and scoring.} Every axis has five severity levels, s1 to s5, ranging from the smallest tested magnitude to a markedly different setup; control frequency is read by rate. A condition is a combination of suite, axis, and level. Table height applies to only two suites, yielding 210 perturbed conditions and four nominal ones. Each condition evaluates its suite's ten tasks from LIBERO's 50 fixed initial states per task, for a total of 500 episodes; LingBot-VA provides a checkpoint only for LIBERO-10. The score is the raw success rate for each axis, averaged over its conditions and compared with the model's unperturbed success rate. The policy receives no fault metadata.

\textbf{Fairness checks.} A perturbed condition provides a valid test of the policy only if the task remains solvable and its goal visible. We verify that every condition has been solved, either a controller with access to the fault parameters or at least one evaluated model, and we audit goal visibility for the camera axes (\appref{sec:supp_diag}).

\textbf{Models.} We evaluate OpenVLA-OFT~\cite{kim2025oft}, OFT-Plus (the LIBERO-Plus authors' checkpoint of the same architecture~\cite{fei2026liberoplus}), \pizero~\cite{black2024pi0}, \pihalf~\cite{pi2025pi05}, X-VLA~\cite{zheng2025xvla}, NORA~\cite{hung2025nora}, MolmoAct~\cite{lee2025molmoact}, and two world-model policies, Cosmos Policy~\cite{kim2026cosmospolicy} and LingBot-VA~\cite{li2026lingbotva} (\appref{sec:supp_models}).

\section{Findings}
\label{sec:findings}

\begin{table}[t]
\centering
\caption{\textbf{Raw success rate per axis, 500 episodes per condition}, Nominal is the unperturbed rate. Bold: best four-suite value per column. $^\dagger$LingBot-VA runs LIBERO-10 only and is not ranked.}
\label{tab:leaderboard}
\footnotesize
\setlength{\tabcolsep}{3pt}
\begin{adjustbox}{max width=\linewidth}
\begin{tabular}{@{}l c ccccccccccc c@{}}
\toprule
Model & Nominal & \rot{\axn{1} cam.\ orient.} & \rot{\axn{2} cam.\ position} & \rot{\axn{3} cam.\ FOV} & \rot{\axn{4} wrist mount} & \rot{\axn{5} frame rot.} & \rot{\axn{6} gain error} & \rot{\axn{7} ctrl.\ freq.} & \rot{\axn{8} proprio.\ noise} & \rot{\axn{9} base pose} & \rot{\axn{10} table height} & \rot{\axn{11} pad geom.} & Mean \\
 & & \rot{\textit{obs.}} & \rot{\textit{obs.}} & \rot{\textit{obs.}} & \rot{\textit{obs.}} & \rot{\textit{latent}} & \rot{\textit{latent}} & \rot{\textit{latent}} & \rot{\textit{latent}} & \rot{\textit{obs.}} & \rot{\textit{obs.}} & \rot{\textit{weak}} & \\
\midrule
$\pi_{0.5}$ & 0.969 & \textbf{0.867} & 0.742 & \textbf{0.929} & 0.674 & 0.751 & 0.932 & \textbf{0.846} & 0.968 & 0.783 & \textbf{0.922} & \textbf{0.889} & \textbf{0.846} \\
Cosmos Policy & 0.973 & 0.801 & 0.774 & 0.862 & \textbf{0.884} & 0.781 & 0.876 & 0.419 & \textbf{0.971} & 0.738 & 0.903 & 0.880 & 0.808 \\
OFT-Plus & 0.970 & \textbf{0.867} & \textbf{0.787} & 0.922 & 0.723 & 0.663 & 0.911 & 0.653 & 0.968 & 0.608 & 0.785 & 0.874 & 0.796 \\
OpenVLA-OFT & 0.966 & 0.701 & 0.590 & 0.874 & 0.730 & 0.673 & 0.910 & 0.661 & 0.964 & 0.589 & 0.803 & 0.872 & 0.761 \\
$\pi_0$ & 0.919 & 0.701 & 0.575 & 0.773 & 0.695 & 0.652 & 0.881 & 0.775 & 0.912 & 0.616 & 0.864 & 0.833 & 0.752 \\
X-VLA & \textbf{0.976} & 0.329 & 0.295 & 0.658 & 0.823 & \textbf{0.968} & \textbf{0.968} & 0.557 & 0.927 & \textbf{0.859} & 0.555 & 0.870 & 0.710 \\
MolmoAct & 0.845 & 0.353 & 0.215 & 0.545 & 0.758 & 0.507 & 0.759 & 0.500 & 0.845 & 0.579 & 0.729 & 0.751 & 0.595 \\
NORA & 0.838 & 0.249 & 0.116 & 0.334 & 0.840 & 0.460 & 0.771 & 0.599 & 0.838 & 0.517 & 0.578 & 0.745 & 0.550 \\
\midrule
LingBot-VA$^\dagger$ & 0.950 & 0.553 & 0.408 & 0.754 & 0.709 & 0.764 & 0.834 & 0.318 & 0.967 & 0.842 & n/a & 0.824 & 0.697$^\dagger$ \\
\bottomrule
\end{tabular}
\end{adjustbox}
\end{table}

\begin{figure}[b]
\centering
\includegraphics[width=0.64\linewidth]{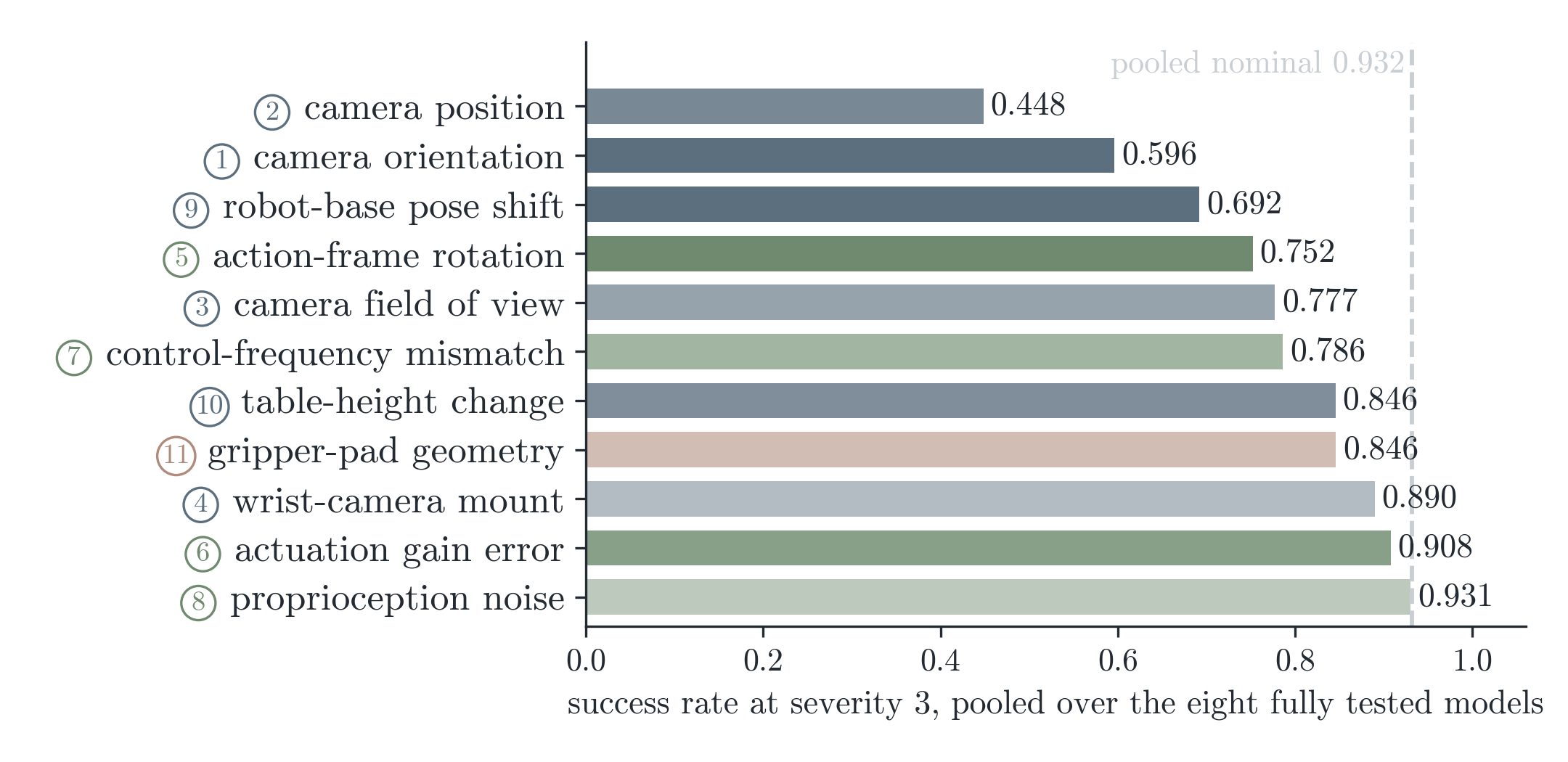}
\caption{\textbf{Success rate by axis at severity 3}, pooled over the eight fully tested models (excluding LingBot-VA); dashed line, pooled nominal; observable axes blue, latent axes green, the weakly observable axis orange, with the axis numbers of \cref{fig:teaser}.}
\label{fig:rq1}
\end{figure}

\cref{tab:leaderboard} presents the main result; pooled results include the eight models evaluated on all four suites, and \appref{sec:supp_grids} reports each model's success rate at every axis and severity level.

\finding{Performance degradation varies across axes.}
At severity 3, mean success rates across the eleven axes range from \axPtwo{} to \axAfive{} (\cref{fig:rq1}; permutation test, $p<0.001$). Moving the scene camera to a new mount cuts success nearly in half; reorienting it in place and shifting the robot's base produce the next-largest drops. At the other end, noise on the reported arm pose causes no observed degradation. The ordering is similar at severities 4 and 5 (\cref{fig:rq1_45}).

\finding{Only the first severity level is free.}
Pooled over models and axes, success moves from \pooledNominal{} to \pooledSa{} at level 1, the only level within a 5 percent non-inferiority margin of nominal (test in \appref{sec:supp_stats}); level 2 reduces success by 9 percent relative to nominal (\pooledSb), with further declines to \pooledSc, \pooledSd, and \pooledSe{} at levels 3, 4, and 5, respectively (\cref{fig:rq2}, \cref{tab:severity}).

\finding{\pihalf ranks first in every suite, but its lead is statistically supported in only three.}
\pihalf ranks first on LIBERO-Spatial, LIBERO-Object, and LIBERO-Goal, remaining first in nearly every resample of each suite's tasks. On LIBERO-10, its lead is narrow: it retains top rank in 65 percent of task resamples with five rivals not statistically distinguishable from it (task-resampled ranks and sign-flip tests; \cref{fig:rq3}, \cref{tab:rq3}, \appref{sec:supp_stats}).

\finding{Robustness is lowest on the long-horizon suite.}
Every model evaluated on all four suites has a lower success rate on LIBERO-10 than its average across the other three suites, with gaps ranging from 0.049 (X-VLA) to 0.184 (NORA) at severity 3. The same pattern holds for every model at severities 4 and 5 (paired sign-flip test, $p=\rqfourP$; \cref{tab:gap}, \cref{fig:rq4_all}).

\finding{Motion errors reduce success, while state-reporting errors do not, suggesting reliance on vision.}
The controller executes each command using the true joint values, while the policy receives the arm state as an input vector; proprioception noise corrupts only the latter, so the arm still moves correctly and success remains unchanged (\axAfive{} versus \pooledNominal{} pooled). Two controls on OpenVLA-OFT support this interpretation: fully corrupting the reported pose leaves success on LIBERO-Spatial at its nominal rate of 0.970, consistent with visual localization of the arm, whereas injecting the same noise into the controller's own state estimate, which introduces actual motion errors, reduces success from 0.960 to 0.470 at 2 cm (\appref{sec:supp_proprio}).

\finding{Visual robustness training does not mitigate faults the camera cannot see; an absolute-pose interface is largely insensitive to calibration faults.}
OFT-Plus, which shares an architecture with OpenVLA-OFT further trained on LIBERO-Plus's perturbations, improves only on the camera axes represented in its training (camera orientation 0.867 versus 0.701, camera position 0.787 versus 0.590) and performs within 0.02 of OpenVLA-OFT on action-frame rotation and every other axis. These results suggest that robustness learned through visual augmentation does not transfer to faults that leave the image unchanged. X-VLA, which outputs absolute poses, shows little degradation under the same calibration faults (0.968 versus a 0.976 nominal), consistent with the hypothesis that an absolute goal allows repeated correction of the remaining pose error, whereas errors in a delta policy can accumulate (\appref{sec:app_more}).

\section{Limitations and Conclusion}
\label{sec:conclusion}

\textbf{Limitations.} All measurements are collected in simulation, using one embodiment, with one axis perturbed per episode. Raw success rates reflect both baseline task proficiency and robustness, and uncertainty intervals are based on ten tasks per suite.

\textbf{Conclusion.} \rig perturbs only the robot rig and leaves every LIBERO task unchanged, allowing us to measure sensitivity to rig changes while holding the task fixed. Nine VLA models succeed on 93 to 98 percent of standard LIBERO episodes; across the eight evaluated on all four suites, pooled success falls to \pooledSc{} at the middle severity and \pooledSe{} at the highest severity. Camera placement is the most damaging axis, calibration and timing faults are not mitigated by visual robustness training, and sensitivity to faults varies with the action interface. The per-axis profiles show practitioners which rig changes each model tolerates (\cref{tab:tolerance}), and we release the benchmark together with manifests specifying the fixed perturbation parameters for each episode.


\bibliography{main}

\clearpage
\appendix
\section{Additional Figures and Tables}
\label{sec:app_main}
\begin{table}[t]
\centering
\caption{\textbf{Cost of each severity level.} Raw success pooled over the eight four-suite models and all axes, with 95 percent intervals from a cluster bootstrap over the 10 tasks of each suite (independent draws per suite). The last columns give the ratio to nominal with its interval and whether the whole interval lies above the 0.95 non-inferiority margin. Only level s1 does.}
\label{tab:severity}
\small
\setlength{\tabcolsep}{4pt}
\begin{adjustbox}{max width=\linewidth}
\begin{tabular}{@{}lcccc@{}}
\toprule
Level & Pooled success & 95\% CI & Ratio to nominal, 95\% CI & Within 5\% margin? \\
\midrule
nominal & 0.932 & [0.913, 0.948] & & \\
s1 & 0.914 & [0.897, 0.929] & 0.981 [0.978, 0.984] & yes \\
s2 & 0.845 & [0.822, 0.865] & 0.907 [0.896, 0.916] & no \\
s3 & 0.767 & [0.742, 0.789] & 0.823 [0.808, 0.836] & no \\
s4 & 0.591 & [0.563, 0.617] & 0.635 [0.614, 0.654] & no \\
s5 & 0.509 & [0.481, 0.536] & 0.546 [0.523, 0.568] & no \\
\bottomrule
\end{tabular}
\end{adjustbox}
\end{table}

\begin{figure}[h]
\centering
\includegraphics[width=\linewidth]{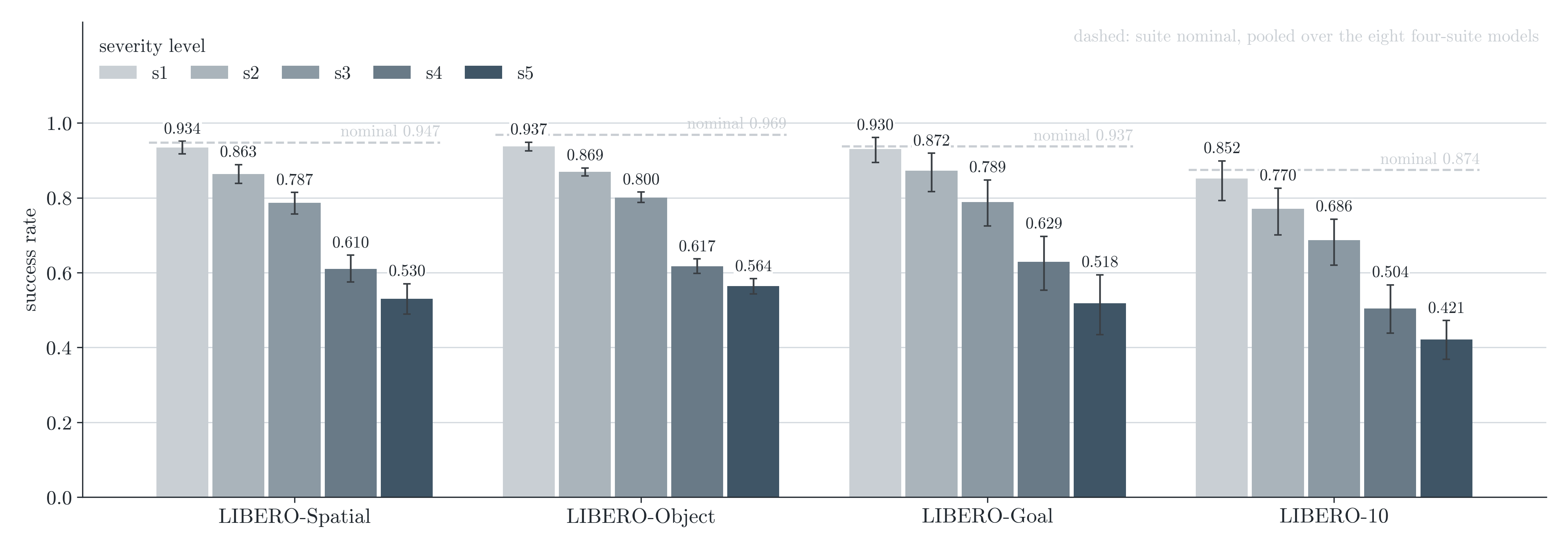}
\caption{\textbf{Success rate by suite and severity level}, level 1 light to level 5 dark, pooled over the eight four-suite models; the dashed line is each suite's pooled nominal.}
\label{fig:rq2}
\end{figure}
\begin{figure}[h]
\centering
\includegraphics[width=0.8\linewidth]{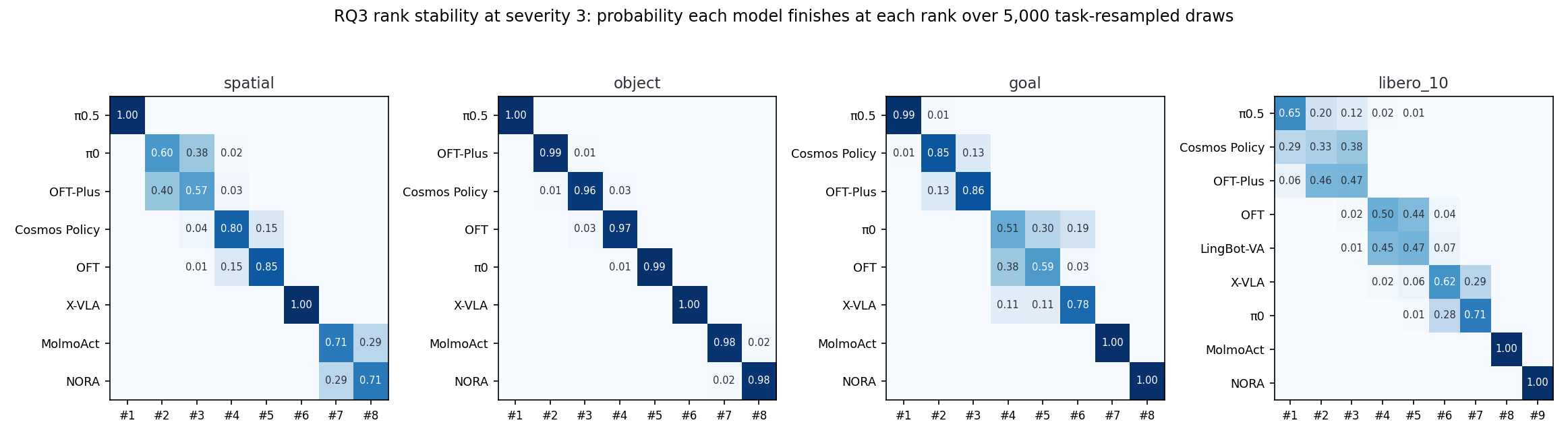}
\caption{\textbf{Rank stability at severity 3.} Probability that each model lands at each rank over 5,000 task-resampled draws, paired across models.}
\label{fig:rq3}
\end{figure}
\begin{table}[t]
\centering
\caption{\textbf{Is the winner certain?} Severity 3. Rank-1 frequency is the fraction of 5,000 task-resampled draws, paired across models within the suite, in which the model ranks first; it describes these 10 tasks. The last column lists the rivals from which Holm-corrected exact paired sign-flip permutation tests over the 10 task differences detect no difference.}
\label{tab:rq3}
\small
\begin{adjustbox}{max width=\linewidth}
\begin{tabular}{@{}llcp{3.6cm}@{}}
\toprule
Suite & Top model & Rank-1 frequency & Rivals not separated (Holm-corrected sign-flip permutation test) \\
\midrule
spatial & $\pi_{0.5}$ & 1.00 & none \\
object & $\pi_{0.5}$ & 1.00 & none \\
goal & $\pi_{0.5}$ & 0.99 & Cosmos Policy \\
LIBERO-10 & $\pi_{0.5}$ & 0.65 & LingBot-VA$^\dagger$, X-VLA, OpenVLA-OFT, OFT-Plus, Cosmos Policy \\
\bottomrule
\end{tabular}
\end{adjustbox}
\end{table}

This appendix gives additional figures and tables (\cref{sec:app_main}), the exact definition and fairness checks of every axis (\cref{sec:supp_axes}), the full results grids and the severity 4 and 5 versions of every analysis (\cref{sec:supp_grids}), the findings moved from the main text (\cref{sec:app_more}), the statistical methods (\cref{sec:supp_stats}), the control-frequency decomposition (\cref{sec:supp_a4}), the proprioception follow-ups (\cref{sec:supp_proprio}), the diagnostics and the scoring decision (\cref{sec:supp_diag}), the cut latency axis (\cref{sec:supp_latency}), model setup and parity (\cref{sec:supp_models}), compute and provenance (\cref{sec:supp_compute}), and the rig-augmentation pilot (\cref{sec:supp_rigaug}).

\section{The Axes in Detail}
\label{sec:supp_axes}
\begin{figure*}[t]
\centering
\includegraphics[width=0.9\linewidth]{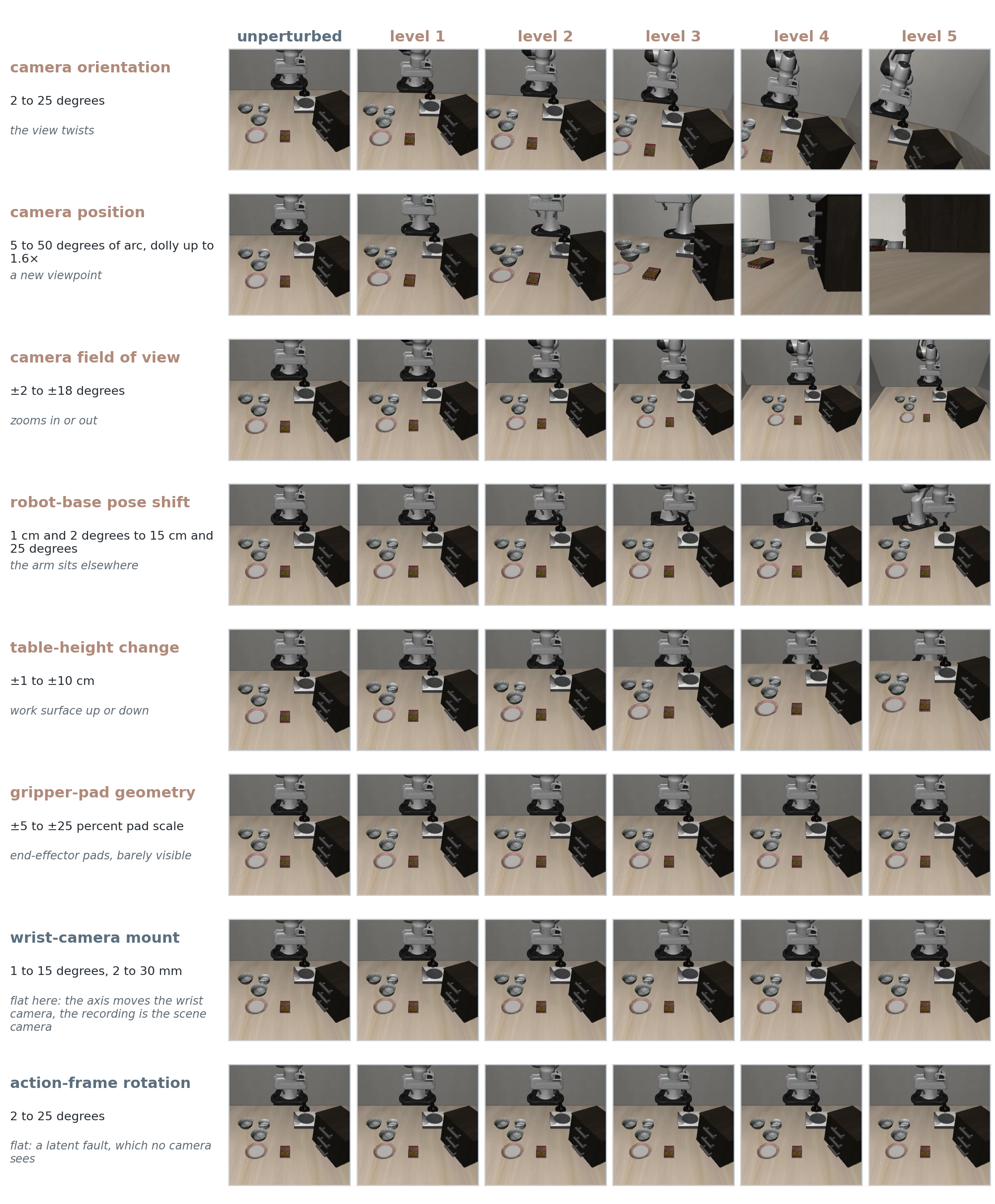}
\caption{\textbf{What the severity ladders look like, axis by axis.} Cosmos Policy, spatial suite, task ``pick up the black bowl between the plate and the ramekin and place it on the plate'', episode 0: the first frame of the same episode under each perturbation, so panels differ only by the rig. Every axis a scene camera can show is included. The action-frame row is flat because the fault is latent. The wrist-camera row is flat only because these renders record the scene camera while that axis moves the wrist camera. On the camera-position row the workspace slides out of shot as the level rises.}
\label{fig:ladders}
\end{figure*}

\cref{fig:ladders} shows the ladders of the observable axes. For every axis we state what physically changes, the exact mathematics of the perturbation, and the fairness checks: the tests that confirm the perturbation preserves task feasibility. An axis that is too weak lets every model pass; one that is too strong breaks the task itself, so zero success no longer indicates policy failure. Four checks apply to every axis and are stated once. A bit-exact parity gate: before any run, the harness must reproduce a fixed set of archived reference results exactly. Frozen manifests: all perturbation values are specified in advance in checksummed files, 107,000 entries, so every episode's perturbation is fixed in advance and replayable under a pinned software configuration. Ladders that start small: levels 1 and 2 are the smaller tested magnitudes (2 and 5 degrees of camera twist, 1 and 2 cm of base displacement), so each ladder brackets the transition instead of starting past it; we do not claim that they represent measured real-world drift. Same tasks, same starts: every axis runs the same 40 tasks from the same fixed initial states as the unperturbed baseline. No check filters the score; all results are published as diagnostics. Notation: $v$ is the ladder value at the condition's severity level; $\pm$ marks a per-episode sign $d$ drawn from the frozen manifest; multi-part axes draw each component's sign independently, so an episode's direction is a sign pattern; magnitudes are fixed within a condition, except that proprioception noise has a fixed standard deviation with a seeded per-episode sample sequence, and control frequency has no direction.

\paragraph{\axnt{1}\ Camera orientation.}
Change: the fixed scene camera's mount is twisted (pitch, yaw, and roll together) by 2 to 25 degrees; its position does not move. Math: in the camera's local frame, pitch, yaw, and roll each receive $\pm v/\sqrt{3}$, applied as successive rotations about the camera's local x, y, and z axes, with the three signs drawn independently per episode (8 patterns); the combined rotation angle equals $v$ to first order and within a few percent at 25 degrees; $v \in \{2, 5, 10, 15, 25\}$ degrees. Fairness checks: the physics is untouched, so the task is always solvable; the risk is the camera turning far enough that goal objects leave the image. The observability audit (\cref{sec:supp_diag}) replays a human demonstration under every perturbed camera and checks, frame by frame with segmentation rendering, that every goal-relevant object stays visible. Episodes where a goal object never appears in the moved camera are published as a list: 5 percent of camera-axis episodes benchmark-wide, nearly all of them out of frame from the first step.

\paragraph{\axnt{2}\ Camera position.}
Change: the camera is remounted somewhere else on a sphere around the workspace (5 to 50 degrees away, and up to 1.6 times farther out or closer in), still aimed at the workspace. Math: on the view sphere centered on the workspace, azimuth and elevation each shift $\pm v/\sqrt{2}$ with independent signs (a combined arc of approximately $v$), and the distance multiplies by $\rho$ or $1/\rho$ (dolly out or in, per-episode sign); the camera is re-aimed at the center; 8 patterns. Ladder $(v, \rho)$: (5$^\circ$, 1.05), (10$^\circ$, 1.10), (20$^\circ$, 1.20), (35$^\circ$, 1.40), (50$^\circ$, 1.60). Fairness checks: the same observability audit. It mattered most here, and it showed that the large collapses are behavioral: OpenVLA-OFT falls to 0.23 on the goal suite at camera position level 4 with no blind episodes at all, so the collapse is the policy breaking, not the task disappearing.

\paragraph{\axnt{3}\ Camera field of view.}
Change: a different lens; the vertical field of view changes by 2 to 18 degrees from the nominal 45 degrees (zoomed in or out). Math: $\mathrm{fovy} = 45^\circ + d\,v$ with per-episode sign $d$; $v \in \{2, 5, 8, 12, 18\}$ degrees. Fairness checks: the observability audit again, since zooming in can push objects out of frame.

\paragraph{\axnt{4}\ Wrist-camera mount.}
Change: the wrist camera's bracket is twisted 1 to 15 degrees and slid 2 to 30 mm along the arm. Math: wrist-camera pitch, yaw, and roll each receive $\pm\theta/\sqrt{3}$, and the mount position shifts $\pm t$ along the bracket axis; all four signs are drawn independently per episode (16 patterns). Ladder $(\theta, t)$: (1$^\circ$, 2 mm), (2$^\circ$, 5 mm), (4$^\circ$, 10 mm), (8$^\circ$, 20 mm), (15$^\circ$, 30 mm). Fairness checks: the observability audit, rendering the perturbed wrist camera itself, finds no blind direction on this axis at any severity (worst-object visibility above 0.1 in every direction). The axis also has a built-in control: OFT-Plus trained with scene-camera randomization but not wrist-camera randomization, and its wrist-camera score is no better than the plain baseline's.

\paragraph{\axnt{5}\ Action-frame rotation.}
Change: every commanded motion executes in a coordinate frame rotated 2 to 25 degrees about vertical (camera-to-arm, or hand-eye, miscalibration). The first image is pixel-identical to the unperturbed run, and later images differ only through the arm's motion. Math: executed action $a' = R(\hat z, d\,v)\,a$, applied to the commanded position delta and rotation delta; the gripper channel passes through; $v \in \{2, 5, 10, 15, 25\}$ degrees. For X-VLA, which is served with absolute-pose actions, the same fault is applied as a rotation of the goal displacement, and the executed trajectories of the absolute and delta forms agree to within $10^{-8}$ in the equivalence test. Fairness checks: compensated replay. The human demonstration is replayed with the exact inverse rotation applied to its commands, $a_{\mathrm{comp}} = R(\hat z, -d\,v)\,a_{\mathrm{demo}}$. Run at levels 3 and 5 on every suite, it succeeds at 93 to 107 percent of the baseline replay rate (the excess over 100 percent is sampling noise), so a controller that knows the fault solves the demonstrated starts at the audited levels, including the harshest.

\paragraph{\axnt{6}\ Actuation gain error.}
Change: all commands execute with excessive or insufficient gain by 5 to 50 percent. Math: $a' = \mathrm{clip}((1 + d\,v)\,a, -1, 1)$ on the six motion channels; $v \in \{0.05, 0.10, 0.20, 0.30, 0.50\}$. The clip matters at level 5, where compensation can saturate. Fairness checks: compensated replay with the inverse $a_{\mathrm{comp}} = a_{\mathrm{demo}}/(1 + d\,v)$ at level 5 on every suite, at 94 to 100 percent of the baseline replay rate. The measured cost is small for most models (0.04 to 0.07 below nominal over the ladder, 0.008 for X-VLA) but not for all: Cosmos Policy loses 0.097 and MolmoAct 0.086, and at level 5 those two fall to 0.618 and 0.550. We hypothesize that re-anchoring each delta on the observed pose keeps magnitude errors from accumulating for most policies, while a 50 percent scale error still breaks grasp-phase precision for some; the clip at $\pm1$ also limits how much a policy could compensate at that level.

\paragraph{\axnt{7}\ Control-frequency mismatch.}
Change: the control loop runs at 25, 10, 50, 5, or 100 Hz instead of the trained 20 Hz. Math: control period $1/f$ with $f \in \{25, 10, 50, 5, 100\}$ Hz, all dividing the 500 Hz physics rate. The episode budget is time-constant: $\mathrm{max\_steps}(f) = \mathrm{max\_steps}(20\,\mathrm{Hz}) \cdot f/20$. On the spatial suite, 220 steps at 20 Hz is 11 seconds; at 5 Hz those 11 seconds contain 55 steps, at 100 Hz 1,100 steps. The controller's per-step motion cap is deliberately not rescaled: how much of each command the servo realizes per period is exactly the fault under test. The severity index is deliberately not monotone in difficulty; interpret each level by its control rate. Fairness checks: our first solvability test, replaying retimed commands open loop, suggested even mild conditions were unsolvable, revealing a limitation of the test. It was replaced with a closed-loop tracking witness, a scripted controller that follows the demonstration's path while re-reading the true arm position every step, evaluated relative to its own unperturbed rate to account for tracking limitations. It certifies 14 of 20 conditions; the goal suite's witness is weak at every severity, a documented witness limitation rather than a task problem. We also pre-registered a prediction that this axis would be mild (hypotheses dated 2026-07-09 in the project record), were wrong, and keep the miss in the record. The mechanism decomposition is in \cref{sec:supp_a4}.

\paragraph{\axnt{8}\ Proprioception noise.}
Change: Gaussian noise (2 to 50 mm position, 0.5 to 8 degrees rotation) on the arm's own sensed pose as reported to the policy. Noise sequences are fixed per episode from the manifest, so runs are reproducible. Math: sensed position $\tilde p = p + \epsilon$ with $\epsilon \sim \mathcal{N}(0, \sigma_p^2 I_3)$; sensed orientation perturbed by a random rotation of standard deviation $\sigma_r$. Ladder $(\sigma_p, \sigma_r)$: (2 mm, 0.5$^\circ$), (5 mm, 1$^\circ$), (10 mm, 2$^\circ$), (20 mm, 4$^\circ$), (50 mm, 8$^\circ$). The gripper state is left clean. Fairness checks: physics untouched, so solvable by construction. The injection path was verified first: the noise is injected into exactly the observation fields the policies consume, so the null result is real, not a wiring gap. Scope, stated precisely: this axis models a fault in the state-reporting path to the policy (miscalibrated forward kinematics, stale or jittery derived state), not encoder failure inside the servo loop. That distinction is now a measurement rather than an argument (\cref{sec:supp_proprio}): injecting the same noise into the controller's own state estimate halves success at 2 cm, where the report-only fault costs nothing. A real encoder bias hits both paths at once, so this axis understates the real fault by design.

\paragraph{\axnt{9}\ Robot-base pose shift.}
Change: the robot's floor mount is moved 1 to 15 cm and rotated 2 to 25 degrees. The objects stay where they are. Math: base position shifts $(\pm\delta/\sqrt{2}, \pm\delta/\sqrt{2}, 0)$ with independent $x$ and $y$ signs (a length-$\delta$ move along one of the four diagonals) and the base yaw shifts $\pm\psi$ independently (8 patterns). Ladder $(\delta, \psi)$: (1 cm, 2$^\circ$), (2 cm, 5$^\circ$), (5 cm, 10$^\circ$), (10 cm, 15$^\circ$), (15 cm, 25$^\circ$). Fairness checks: the closed-loop tracking witness, driving to the same world positions from the new base: 26 of 40 direction branches certified. Our audit of the rejected branches found 13 of 14 solvable anyway, because evaluated models solve them, which is part of why certification informs but no longer filters.

\paragraph{\axnt{10}\ Table-height change.}
Change: the work surface is raised or lowered 1 to 10 cm; the objects' starting positions move with the table. Only two suites have height-adjustable arenas. Math: the table height shifts by $d\,v$, and every free object's starting height in the paired initial state shifts by the same amount; $v \in \{1, 2, 4, 7, 10\}$ cm. The axis is registered only where every task in the suite supports the shift (spatial and goal); the wrapper checks the arena type at every reset and raises an error if the shift would be ignored, so an episode scored as perturbed but actually running nominal cannot exist. Quantified caveat: restricting any of the ten all-suite axes to the same two suites shifts its mean by $-0.05$ to $+0.12$ depending on the axis and the model, so the table-height value uses a different suite mix from the other axes; read it against the two-suite nominal as in \cref{sec:benchmark}. Fairness checks: the premise was verified in the simulator model first (the robot is floor-mounted, not table-mounted, so a moving table genuinely changes the geometry between arm and objects), then the same tracking witness as the base-pose axis: 10 of 20 branches certified, with rejected branches mostly solvable by models.

\paragraph{\axnt{11}\ Gripper-pad geometry.}
Change: the two rubber pads on the end effector are scaled 5 to 25 percent in all three dimensions (worn down or oversized). Fingers, joints, and travel are untouched. Math: pad half-extents scale by $(1 + d\,v)$ from the nominal $8 \times 4 \times 8$ mm, with the pad center shifted so that the face mounted on the finger stays fixed; $v \in \{0.05, 0.10, 0.15, 0.20, 0.25\}$. Effect on the opening: 2 mm lost at $+25$ percent, out of 79 mm, which is why opening width does not explain the degradation. Fairness checks: the deepest-audited axis, because ``does gripping become impossible?'' is exactly the wrong-question risk. Four tests: demo-replay survival at every level, published as context and never as a filter (a replayed demonstration cannot adapt its grasp, so failing it proves nothing); a width oracle showing that no object grasp is blocked by the opening width even at the largest pads; a per-direction split showing that worn pads are free everywhere (replay survival at 89 to 100 percent of the baseline replay rate at every level) while oversized pads degrade gradually (to 37 to 72 percent of baseline at the worst level), so the damage is adaptation difficulty, not impossibility; and for the one task that every model fails or nearly fails at the largest growth (LIBERO-10's ``put the yellow and white mug in the microwave and close it'', where no model succeeds on more than 2 of its 22 oversized-pad episodes), a contact-level investigation showing that the grasp still succeeds but the mug is dropped mid-transport as the wider pad degrades the hold. This indicates grasp instability rather than task infeasibility.

\section{Full Results and the Severity 4 and 5 Analyses}
\label{sec:supp_grids}
\begin{figure*}[t]
\centering
\includegraphics[width=\linewidth]{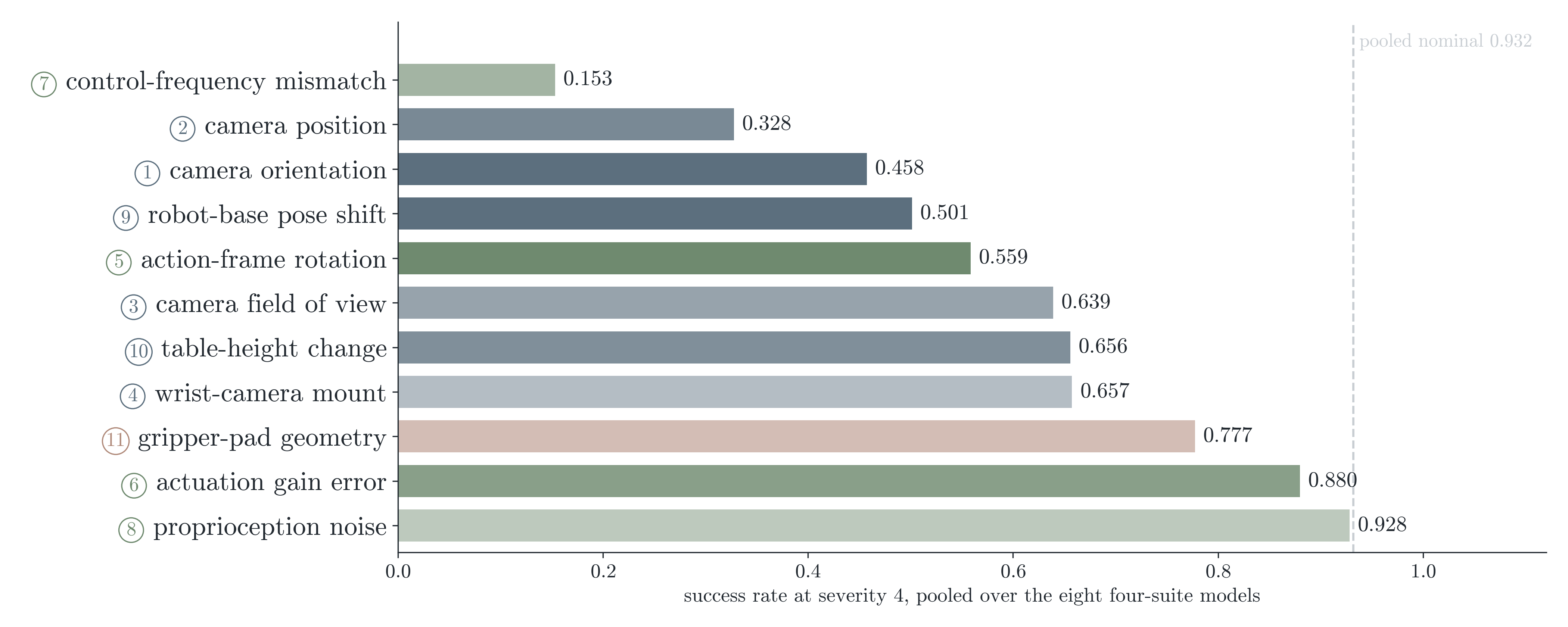}\\[4pt]
\includegraphics[width=\linewidth]{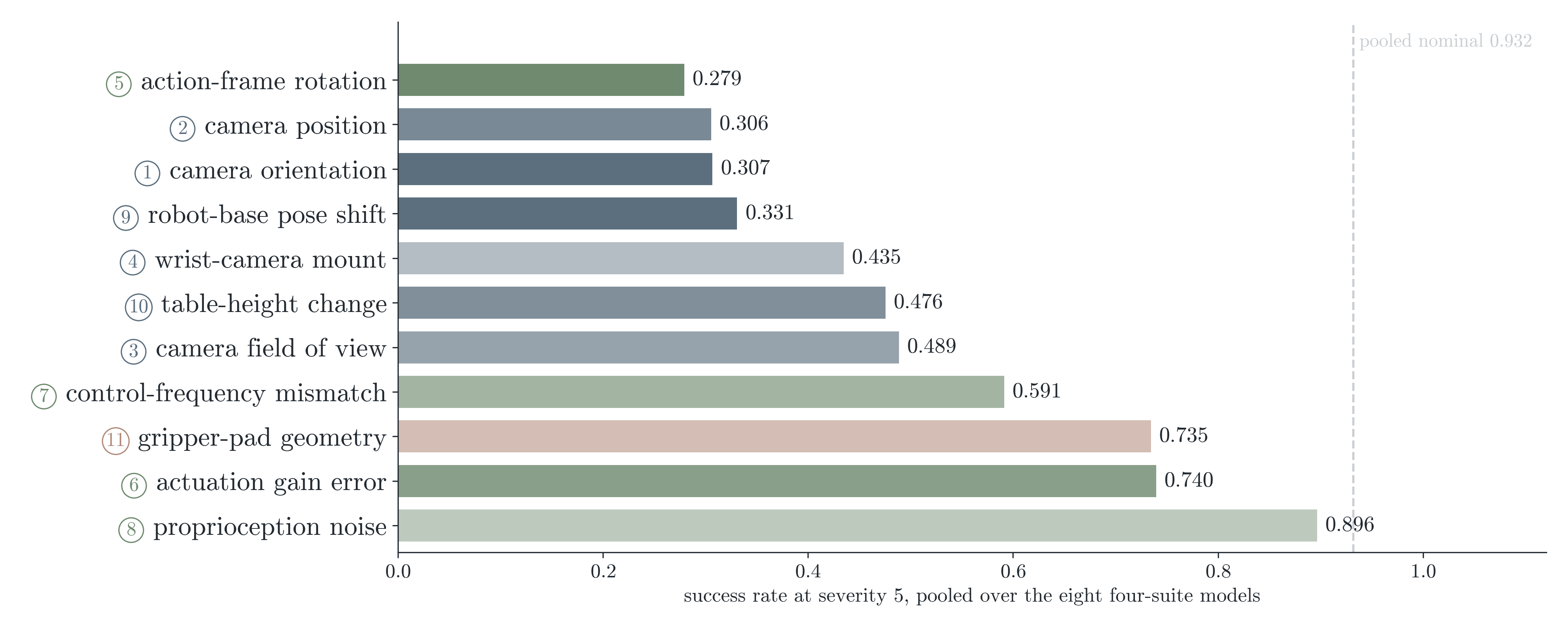}
\caption{\textbf{Success rate by axis at severities 4 (top) and 5 (bottom)}, pooled over the eight four-suite models; dashed line is the pooled nominal; colors as in \cref{fig:rq1}.}
\label{fig:rq1_45}
\end{figure*}

\begin{figure*}[t]
\centering
\includegraphics[width=\linewidth]{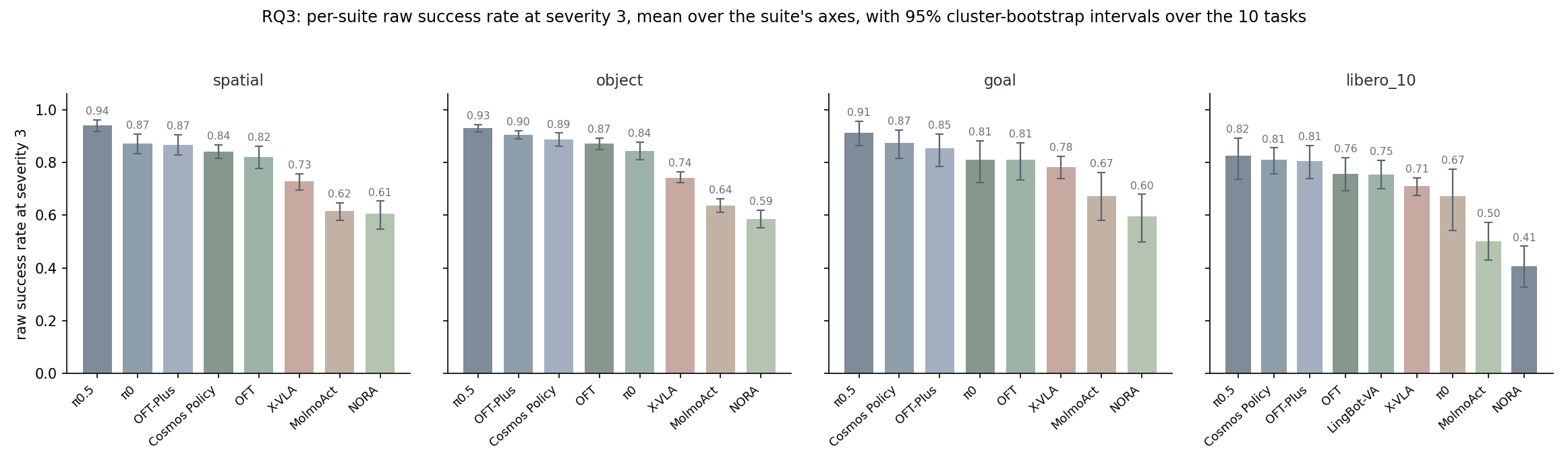}\\[4pt]
\includegraphics[width=\linewidth]{figures/rq3_rank_s3.png}
\caption{\textbf{Per-suite success with intervals (top) and rank stability (bottom) at severity 3.} Top: raw success rate per suite, mean over the suite's axes, with 95 percent intervals from a cluster bootstrap over the 10 tasks, the same task resamples for every model. Bottom: as \cref{fig:rq3}.}
\label{fig:rq3_45}
\end{figure*}

\begin{figure*}[t]
\centering
\includegraphics[width=\linewidth]{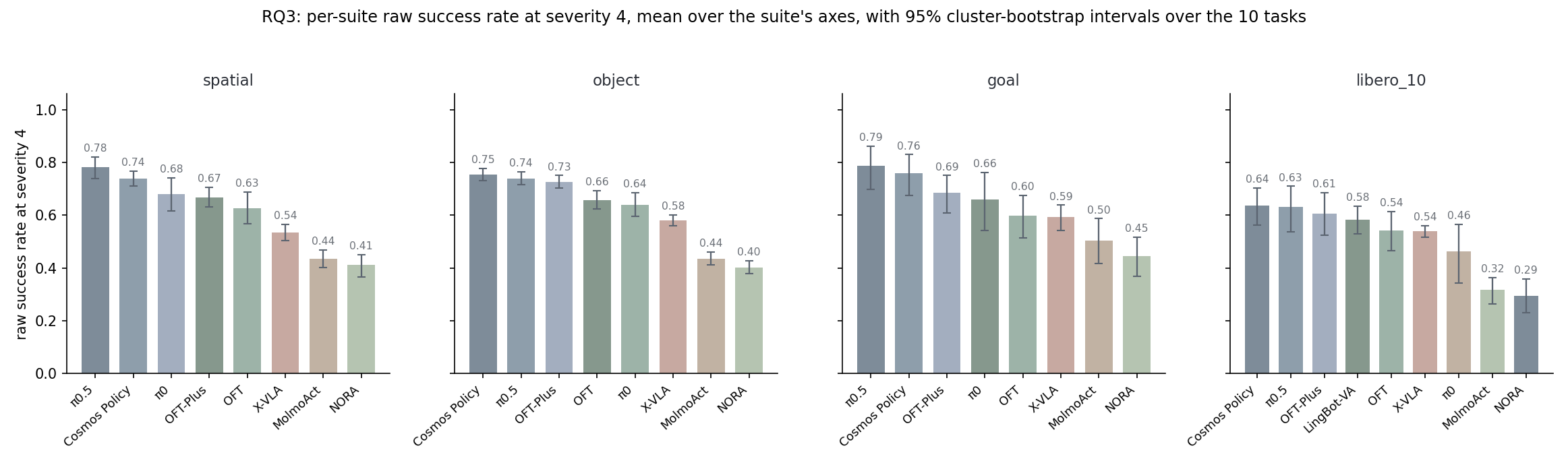}\\[4pt]
\includegraphics[width=\linewidth]{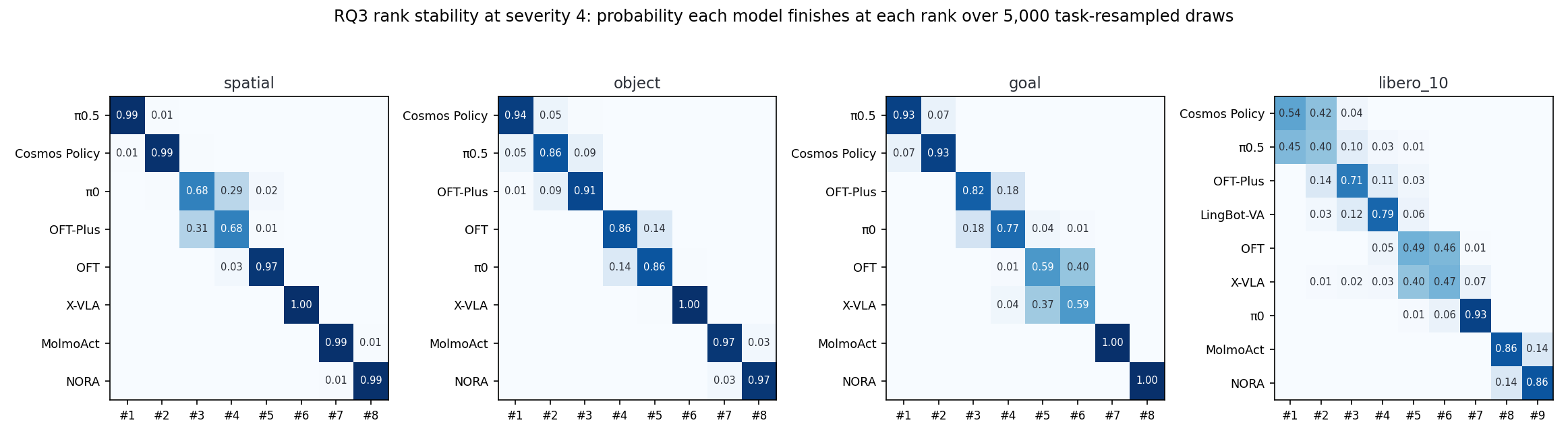}
\caption{\textbf{Per-suite success with intervals (top) and rank stability (bottom) at severity 4}, as in \cref{fig:rq3_45}. Cosmos Policy takes the object and LIBERO-10 suites.}
\label{fig:rq3_45b}
\end{figure*}

\begin{figure*}[t]
\centering
\includegraphics[width=\linewidth]{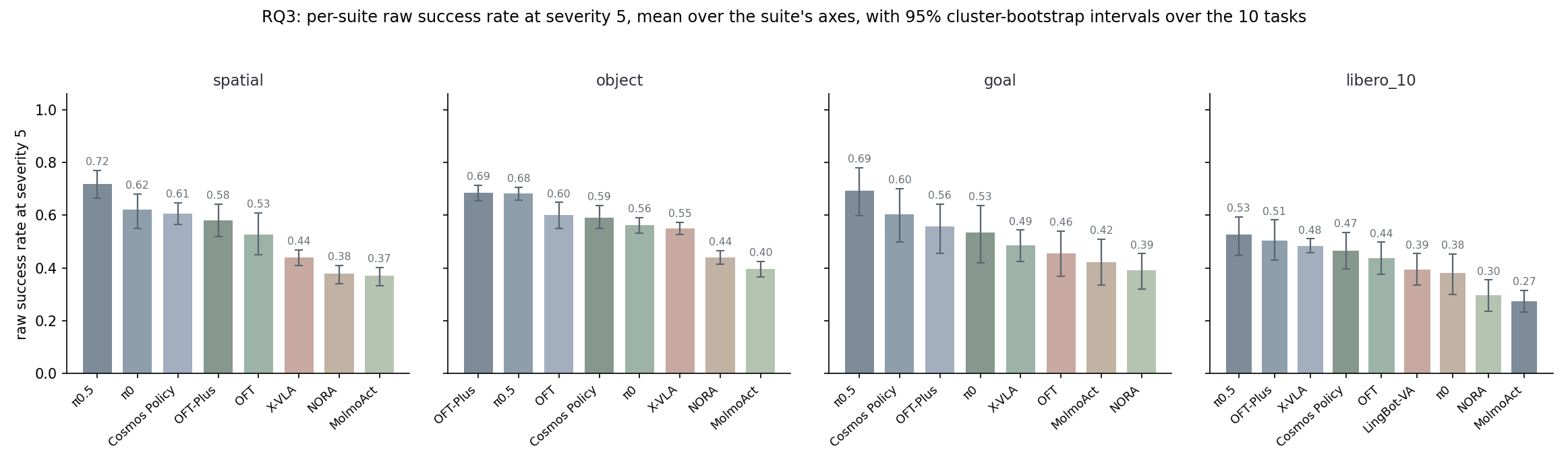}\\[4pt]
\includegraphics[width=\linewidth]{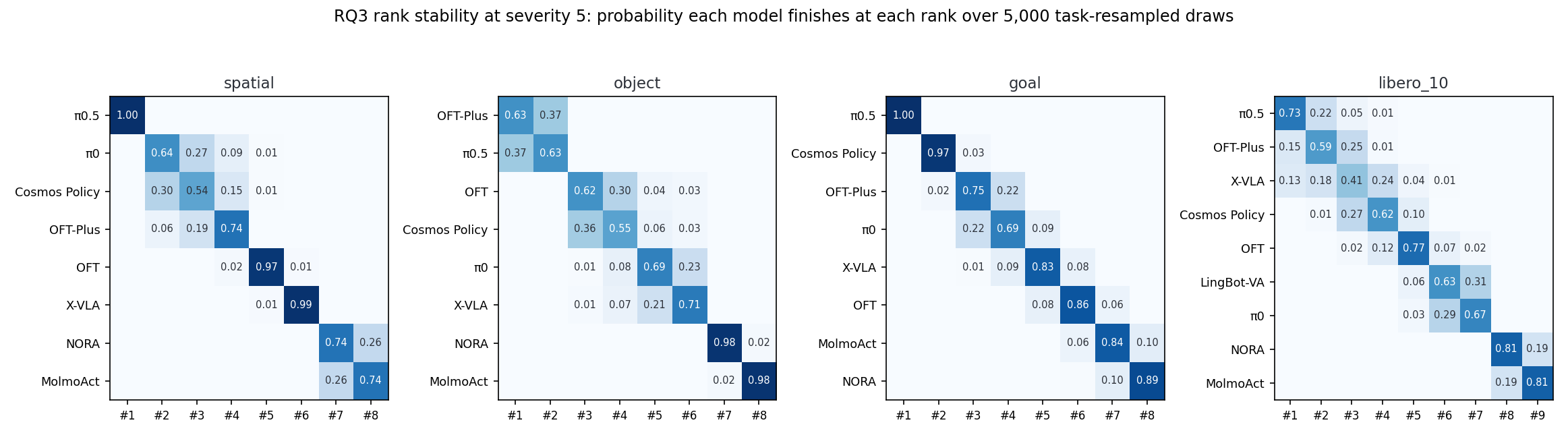}
\caption{\textbf{Per-suite success with intervals (top) and rank stability (bottom) at severity 5}, as in \cref{fig:rq3_45}. First place in the object suite is a coin flip between OFT-Plus and \pihalf.}
\label{fig:rq3_45c}
\end{figure*}

\begin{figure*}[t]
\centering
\includegraphics[width=\linewidth]{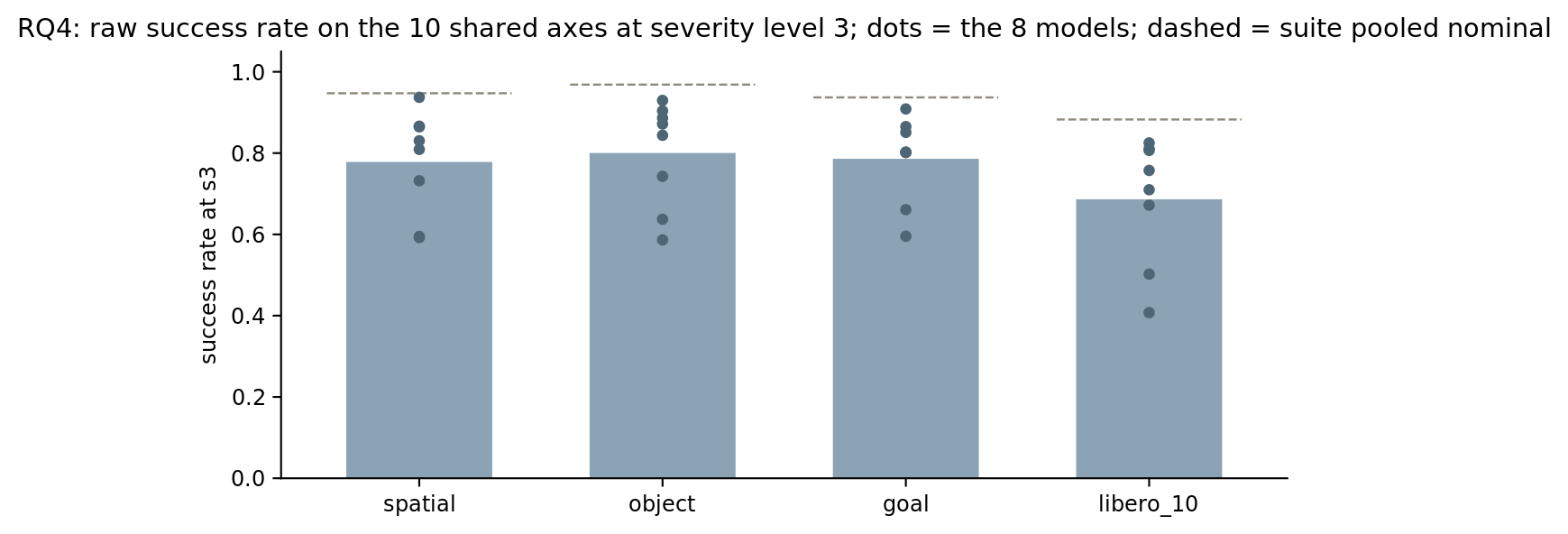}\\[4pt]
\includegraphics[width=\linewidth]{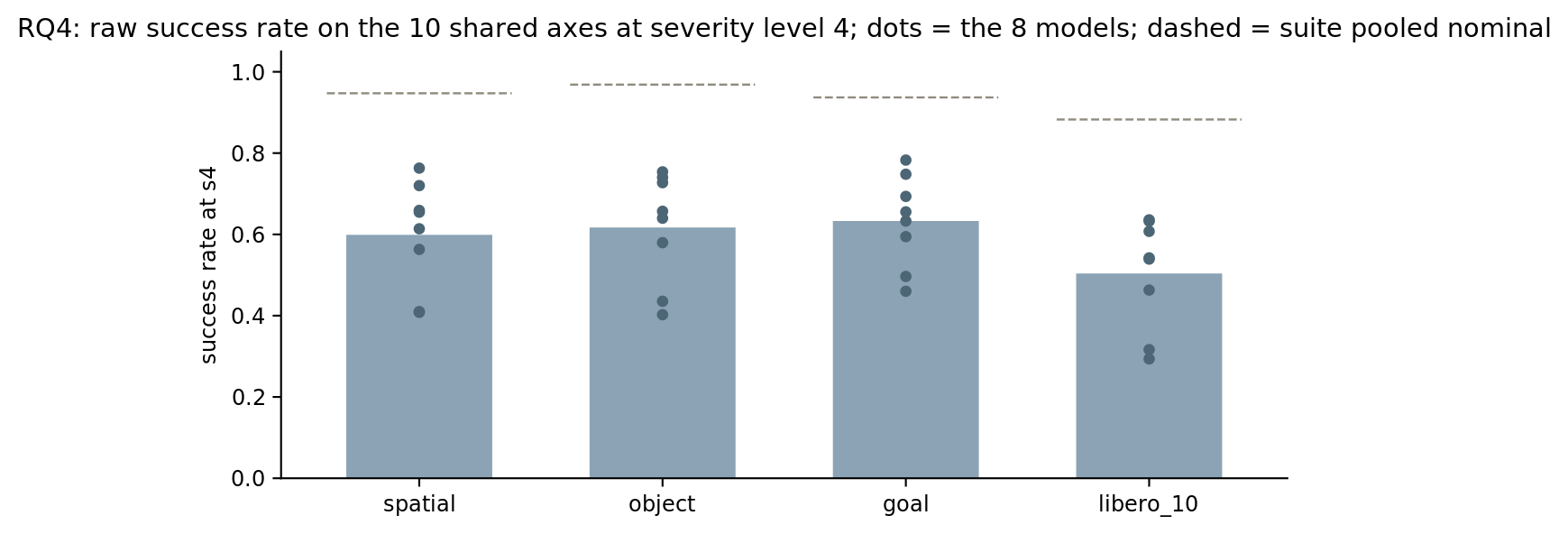}\\[4pt]
\includegraphics[width=\linewidth]{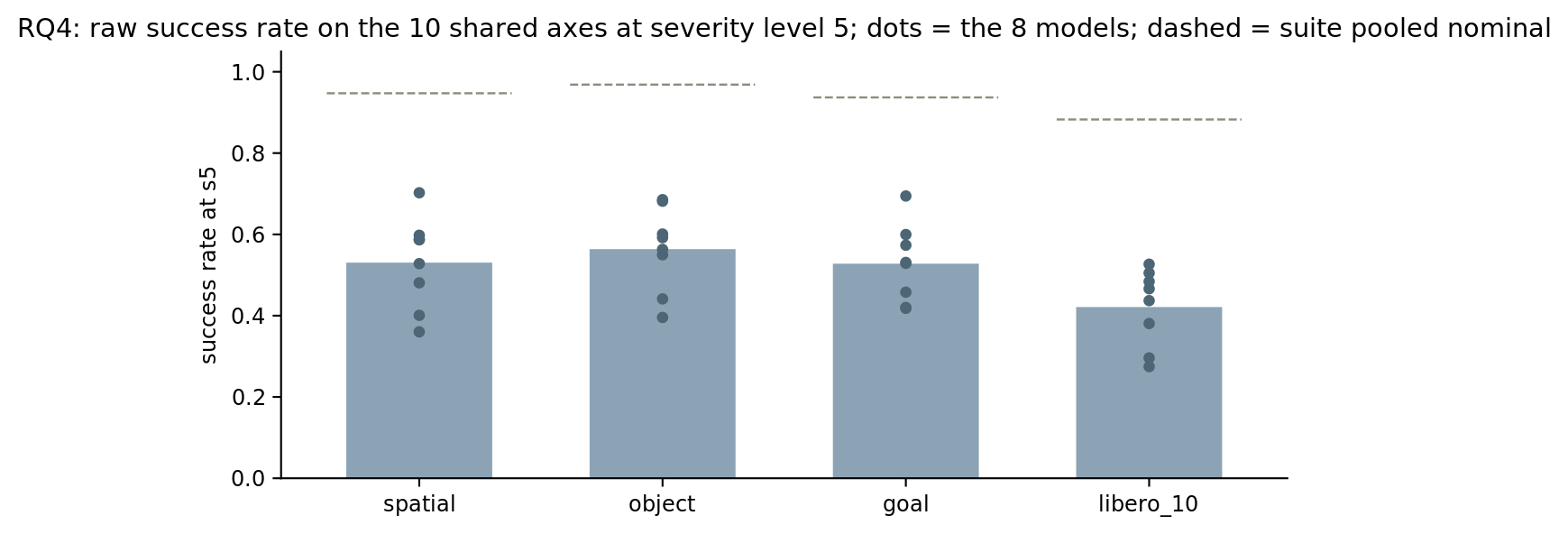}
\caption{\textbf{Success rate by suite on the ten shared axes} at severities 3, 4, and 5 (top to bottom). Dots are the eight four-suite models.}
\label{fig:rq4_all}
\end{figure*}

\begin{figure*}[t]
\centering
\includegraphics[width=0.72\linewidth]{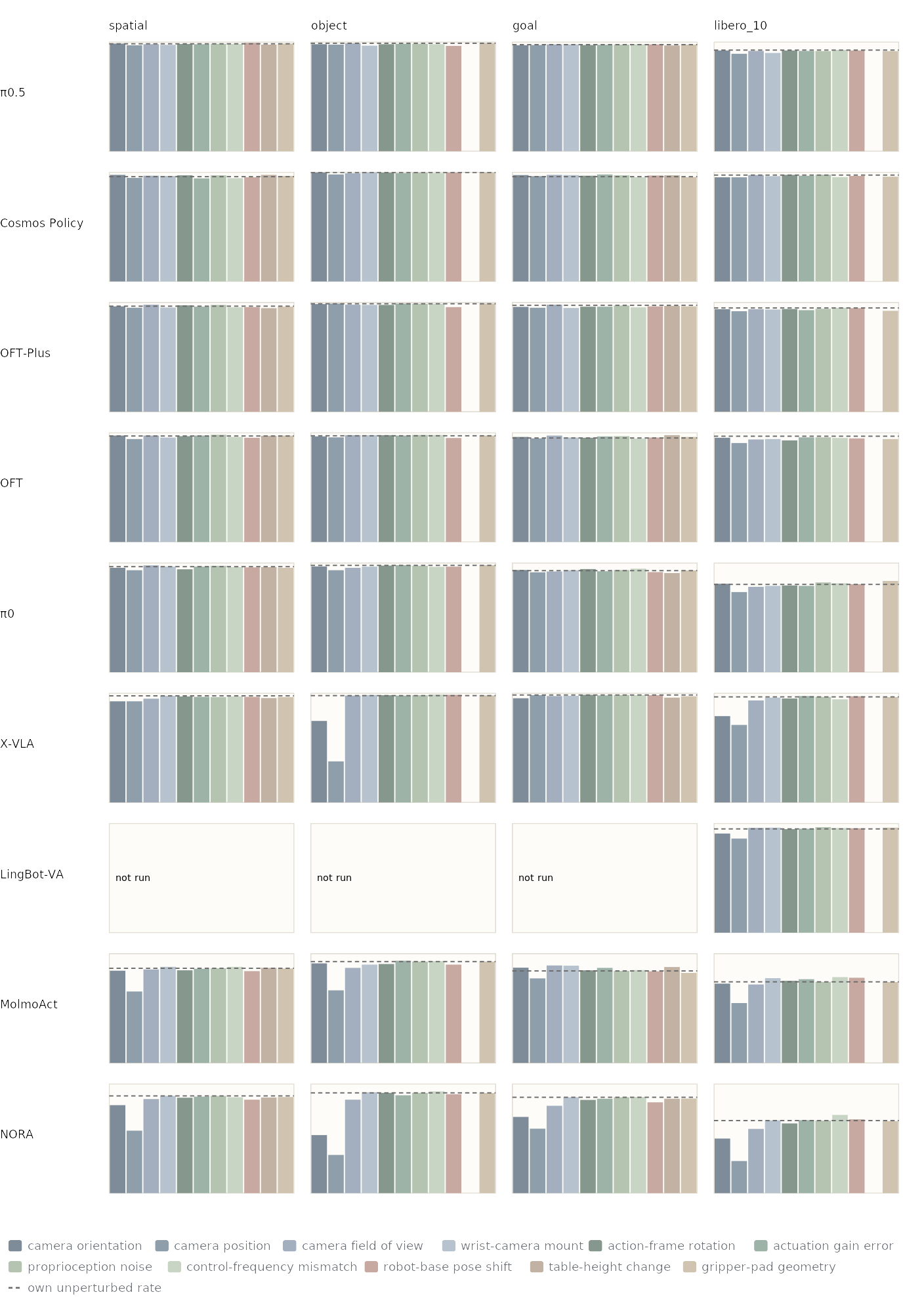}
\caption{\textbf{Full results grid, severity 1.} Rows are models, columns the four suites, 11 bars per panel in a fixed axis order. The dashed line is that model's own unperturbed rate on that suite, so a bar reaching the line means the perturbation cost nothing. The numbers are in \cref{tab:grid1,tab:grid2,tab:grid3,tab:grid4,tab:grid5}. This figure plots \cref{tab:grid1}; \cref{fig:grid_s2,fig:grid_s3,fig:grid_s4,fig:grid_s5} plot \cref{tab:grid2,tab:grid3,tab:grid4,tab:grid5} for severities 2 to 5.}
\label{fig:grids}\label{fig:grid_s1}
\end{figure*}

\cref{tab:grid1,tab:grid2,tab:grid3,tab:grid4,tab:grid5} give the raw success rate of every model on every suite and axis at each severity level, with the model's unperturbed rate on that suite beside it; \cref{fig:grids} plots the same numbers. \cref{fig:rq1_45,fig:rq3_45,fig:rq3_45b,fig:rq3_45c,fig:rq4_all} repeat the analyses of \cref{sec:findings} at severities 4 and 5, and \cref{tab:crosscheck} summarizes how the four benchmark-level findings change with the level. The main findings persist. At level 4 the axis spread is largest (0.775) and control-frequency mismatch becomes the worst axis (\ctrlFiveHz, the 5 Hz rung); at level 5 floor effects compress the axes to \letterGroupsSfive{} letter groups and action-frame rotation (0.279) edges out camera position (0.306), camera orientation (0.307), and base pose (0.331) at the bottom. \cref{tab:gap} lists the per-model long-horizon gaps: all eight are negative at every level (exact Wilcoxon $p = 0.008$ at levels 3, 4, and 5).

\cref{tab:tolerance} reads the same grids as a tolerance per axis: the harshest level each model passes before its first failure, under an absolute threshold of 0.80 and a threshold of 0.90 of the model's own nominal rate. Under the absolute bar, every model passes wrist-camera mount, actuation gain, and proprioception noise at level 3, while three pass camera orientation, action-frame rotation, and base pose, and one passes camera position; at level 5 no model keeps camera orientation, camera position, base pose, or table height, and seven of eight keep proprioception noise.

\begin{table*}[t]
\centering
\caption{\textbf{Tolerance per axis.} The harshest severity level each model passes before its first failure on the axis, given as that level's leading parameter (ladders in \cref{tab:axes}, both parameters of every level in \appref{sec:supp_axes}), read from the pooled success per level (\cref{tab:grid1,tab:grid2,tab:grid3,tab:grid4,tab:grid5}). Criterion A passes a level when success is at least 0.80; criterion B when it is at least 0.90 of the model's own nominal rate. $\geq$ marks a model that passes the harshest level, $<$ one that fails the mildest. Control frequency \axn{7} is not ordered by dose, so its cell gives the range of rates passed around the nominal 20 Hz. An asterisk marks a cell whose deciding rate lies within 0.02 of the bar. Levels are set per axis, so the table compares models within an axis, not axes with one another. $^\dagger$LingBot-VA is evaluated on LIBERO-10 only, where table height is not applied.}
\label{tab:tolerance}
\small
\setlength{\tabcolsep}{4pt}
\begin{adjustbox}{max width=\textwidth}
\begin{tabular}{@{}l c ccccccccccc@{}}
\toprule
Model & Nominal & \rot{\axn{1} cam.\ orient.\ ($^\circ$)} & \rot{\axn{2} cam.\ position ($^\circ$ arc)} & \rot{\axn{3} cam.\ FOV ($\pm^\circ$)} & \rot{\axn{4} wrist mount ($^\circ$)} & \rot{\axn{5} frame rot.\ ($^\circ$)} & \rot{\axn{6} gain error ($\pm$\%)} & \rot{\axn{7} ctrl.\ freq.\ (Hz)} & \rot{\axn{8} proprio.\ noise (mm)} & \rot{\axn{9} base pose (cm)} & \rot{\axn{10} table height ($\pm$cm)} & \rot{\axn{11} pad geom.\ ($\pm$\%)} \\
\midrule
\multicolumn{13}{@{}l}{\textit{Criterion A: success at least 0.80}} \\
$\pi_{0.5}$ & 0.969 & 15 & 10 & $\geq$18 & 4 & 10 & $\geq$50 & 10--100 & $\geq$50 & 5 & 7$^{*}$ & $\geq$25$^{*}$ \\
Cosmos Policy & 0.973 & 10 & 10 & 12 & 8 & 10 & 30 & 25 & $\geq$50 & 5$^{*}$ & 7 & 20$^{*}$ \\
OFT-Plus & 0.970 & 15 & 20 & $\geq$18 & 4 & 5 & 30 & 25--50 & $\geq$50 & 2 & 4 & 20$^{*}$ \\
OpenVLA-OFT & 0.966 & 5 & 10 & 12 & 4 & 5$^{*}$ & 30 & 25--50 & $\geq$50 & 2$^{*}$ & 4 & 20$^{*}$ \\
$\pi_0$ & 0.919 & 5 & 5 & 8 & 4 & 5 & 30 & 10--100 & $\geq$50 & 2 & 7$^{*}$ & 15 \\
X-VLA & 0.976 & 2 & $<$5 & 5 & 4 & $\geq$25 & $\geq$50 & 10--25$^{*}$ & 20 & 10$^{*}$ & 2 & 15$^{*}$ \\
MolmoAct & 0.845 & 2 & $<$5 & 2 & 4 & 2 & 20$^{*}$ & 25 & $\geq$50 & 1 & 4$^{*}$ & 10$^{*}$ \\
NORA & 0.838 & $<$2 & $<$5 & $<$2 & $\geq$15 & 2 & 20$^{*}$ & 25$^{*}$ & $\geq$50 & 1$^{*}$ & 2$^{*}$ & 10$^{*}$ \\
LingBot-VA$^\dagger$ & 0.950 & 2 & 5 & 8 & 4 & 10 & 30 & 25 & $\geq$50 & 10$^{*}$ & -- & 15$^{*}$ \\
\midrule
\multicolumn{13}{@{}l}{\textit{Criterion B: success at least 0.90 of the model's own nominal rate}} \\
$\pi_{0.5}$ & 0.969 & 10$^{*}$ & 10 & 12 & 4 & 5 & 30 & 10--100 & $\geq$50 & 2$^{*}$ & 7 & 15 \\
Cosmos Policy & 0.973 & 5$^{*}$ & 10 & 8 & 8 & 10$^{*}$ & 30 & 25 & $\geq$50 & 2 & 7 & 15$^{*}$ \\
OFT-Plus & 0.970 & 10 & 10 & 12 & 2$^{*}$ & 5 & 30 & 25 & $\geq$50 & 1 & 4$^{*}$ & 15$^{*}$ \\
OpenVLA-OFT & 0.966 & 5 & 5$^{*}$ & 8 & 4$^{*}$ & 5 & 30 & 25 & $\geq$50 & 1 & 4 & 15$^{*}$ \\
$\pi_0$ & 0.919 & 5 & 5 & 8$^{*}$ & 4 & 5 & 30 & 25--100$^{*}$ & $\geq$50 & 2 & 4$^{*}$ & 15$^{*}$ \\
X-VLA & 0.976 & $<$2 & $<$5 & 2$^{*}$ & 4 & $\geq$25 & $\geq$50 & 25 & 20 & 5 & 2 & 10$^{*}$ \\
MolmoAct & 0.845 & 2 & $<$5 & 2$^{*}$ & 4 & 5$^{*}$ & 20$^{*}$ & 25 & $\geq$50 & 2$^{*}$ & 4 & 15$^{*}$ \\
NORA & 0.838 & $<$2 & $<$5 & 2 & $\geq$15 & 2$^{*}$ & 30$^{*}$ & 25--50 & $\geq$50 & 1$^{*}$ & 2 & 10$^{*}$ \\
LingBot-VA$^\dagger$ & 0.950 & 2 & 5$^{*}$ & 5 & 4$^{*}$ & 10 & 20$^{*}$ & 25 & $\geq$50 & 5 & -- & 10 \\
\bottomrule
\end{tabular}

\end{adjustbox}
\end{table*}

\begin{table}[t]
\centering
\caption{\textbf{The four benchmark-level findings at severities 3, 4, and 5.} Mean axis success is the mean of the 11 axis means (the episode-pooled rates are \pooledSc, \pooledSd, and \pooledSe). The letter-group count at level 4 is seed dependent (5 or 6) because two pairs sit at the corrected 0.05 threshold. P(rank 1) is listed for spatial / object / goal / LIBERO-10.}
\label{tab:crosscheck}
\small
\setlength{\tabcolsep}{3.5pt}
\begin{adjustbox}{max width=\linewidth}\begin{tabular}{@{}lccc@{}}
\toprule
Criterion & s3 & s4 & s5 \\
\midrule
mean axis success (nominal \pooledNominal) & 0.770 & 0.594 & 0.508 \\
axis spread, max minus min & 0.482 & 0.775 & 0.617 \\
letter groups the axes resolve into & 5 & 5 or 6 & 4 \\
P(rank 1) of the suite winners & 1.00 / 1.00 / 0.99 / 0.65 & 1.00 / 0.93 / 0.93 / 0.53 & 1.00 / 0.62 / 1.00 / 0.73 \\
Holm-significant top-vs-rest pairs (of 29) & 23 & 22 & 24 \\
long-horizon gap, exact Wilcoxon $p$ & 0.008 & 0.008 & 0.008 \\
\bottomrule
\end{tabular}\end{adjustbox}
\end{table}

\begin{table}[t]
\centering
\caption{\textbf{Long-horizon gap per model at severity 3.} Success on LIBERO-10 minus the mean over the other three suites, over the ten axes shared by all suites (table height excluded).}
\label{tab:gap}
\small
\begin{tabular}{@{}lc@{}}
\toprule
Model & libero-10 minus mean of other suites (s3) \\
\midrule
NORA & $-0.184$ \\
$\pi_0$ & $-0.165$ \\
MolmoAct & $-0.129$ \\
$\pi_{0.5}$ & $-0.100$ \\
OpenVLA-OFT & $-0.070$ \\
OFT-Plus & $-0.067$ \\
Cosmos Policy & $-0.050$ \\
X-VLA & $-0.049$ \\
\bottomrule
\end{tabular}

\end{table}


\begin{figure*}[t]
\centering
\includegraphics[width=0.72\linewidth]{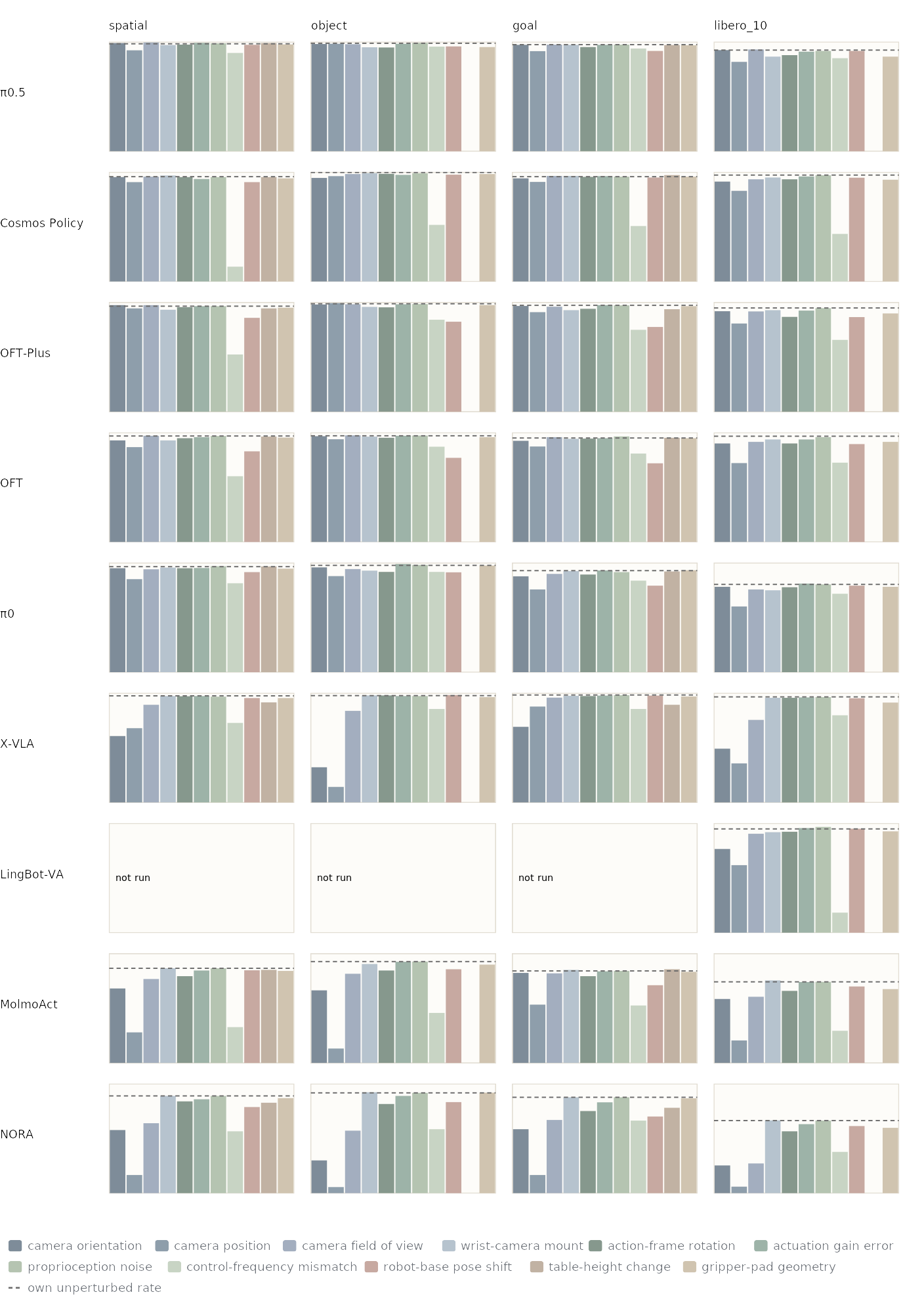}
\caption{\textbf{Full results grid, severity 2}, as in \cref{fig:grids}; the numbers are in \cref{tab:grid2}.}
\label{fig:grid_s2}
\end{figure*}
\begin{figure*}[t]
\centering
\includegraphics[width=0.72\linewidth]{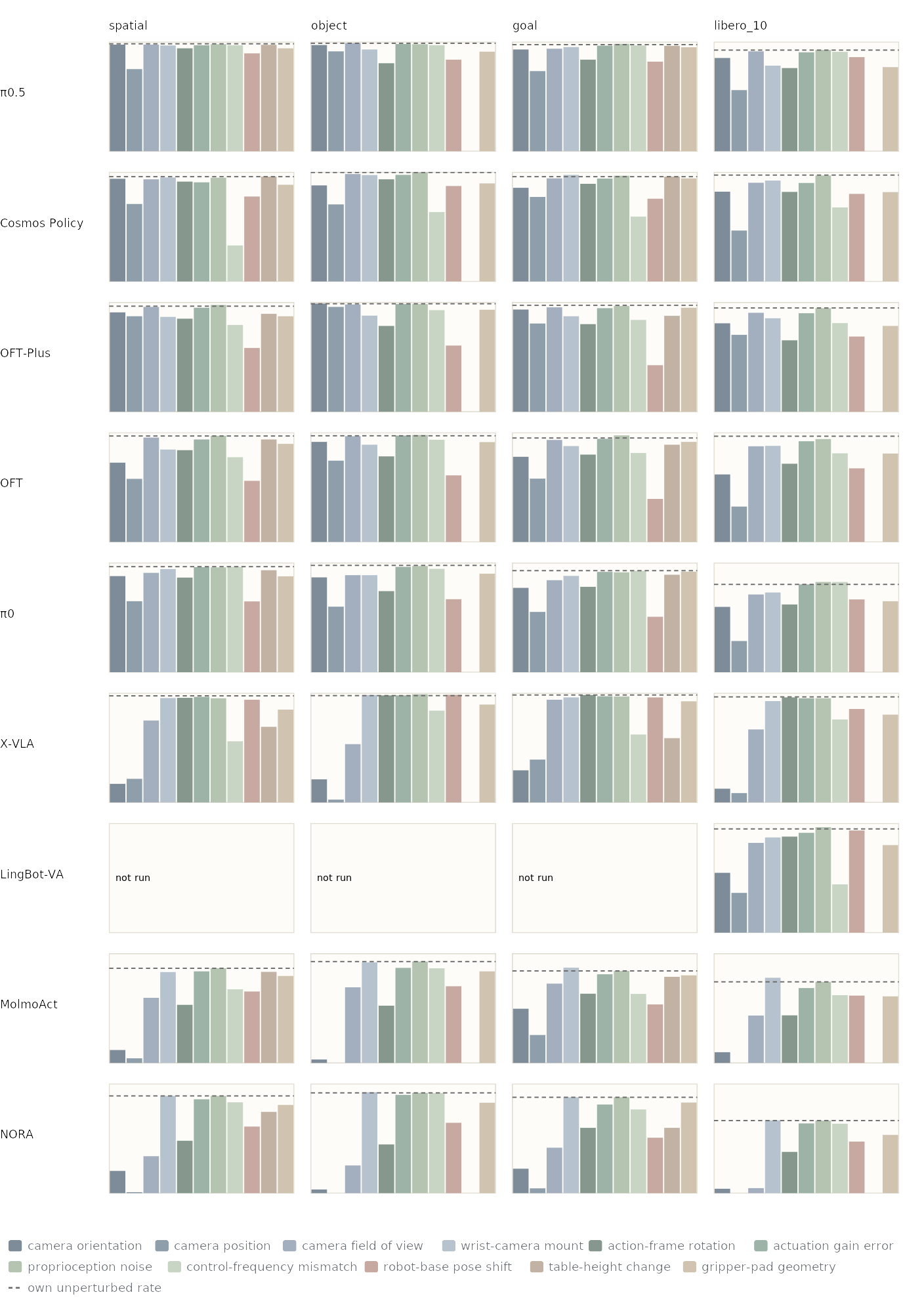}
\caption{\textbf{Full results grid, severity 3}, as in \cref{fig:grids}; the numbers are in \cref{tab:grid3}.}
\label{fig:grid_s3}
\end{figure*}
\begin{figure*}[t]
\centering
\includegraphics[width=0.72\linewidth]{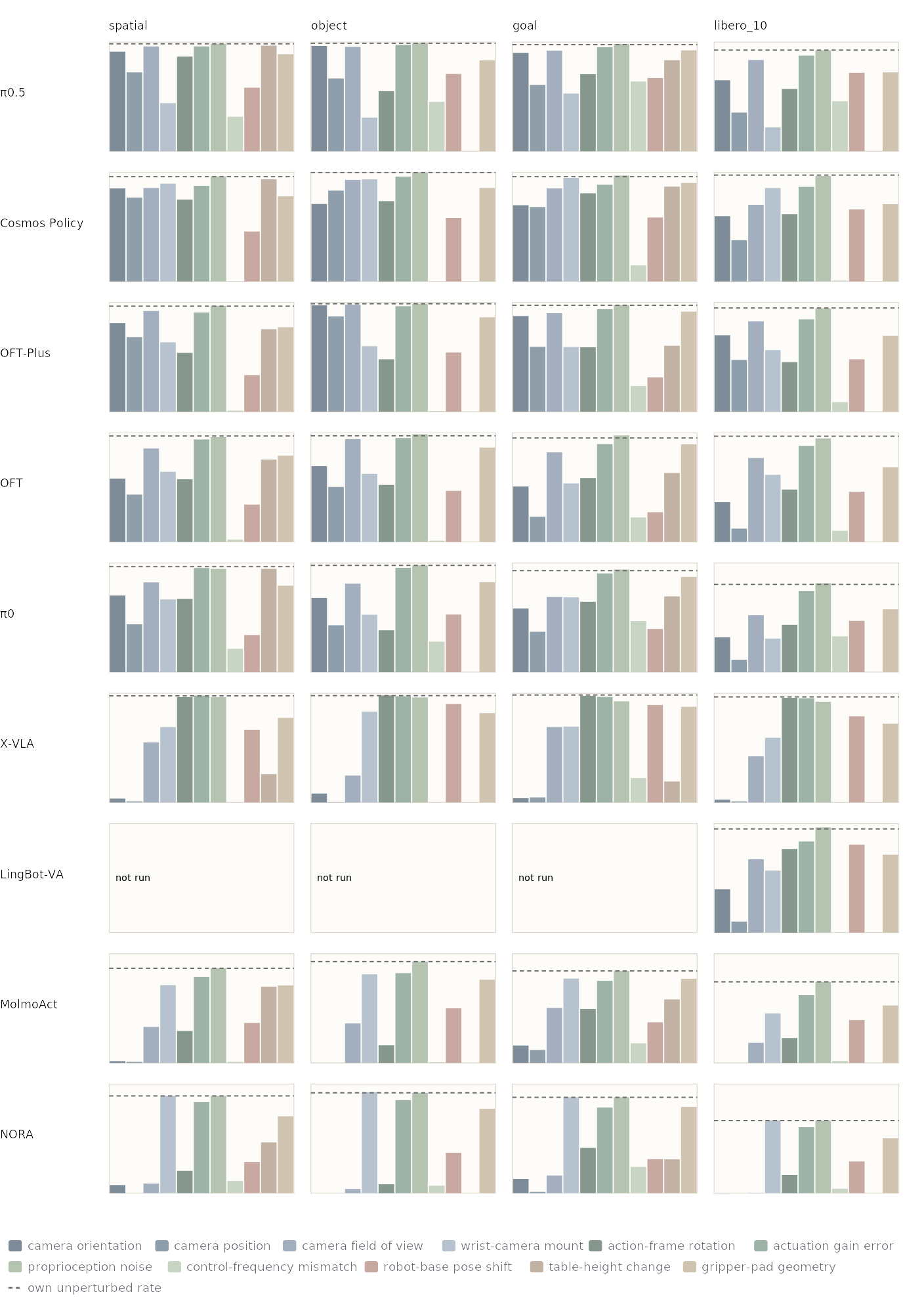}
\caption{\textbf{Full results grid, severity 4}, as in \cref{fig:grids}; the numbers are in \cref{tab:grid4}.}
\label{fig:grid_s4}
\end{figure*}
\begin{figure*}[t]
\centering
\includegraphics[width=0.72\linewidth]{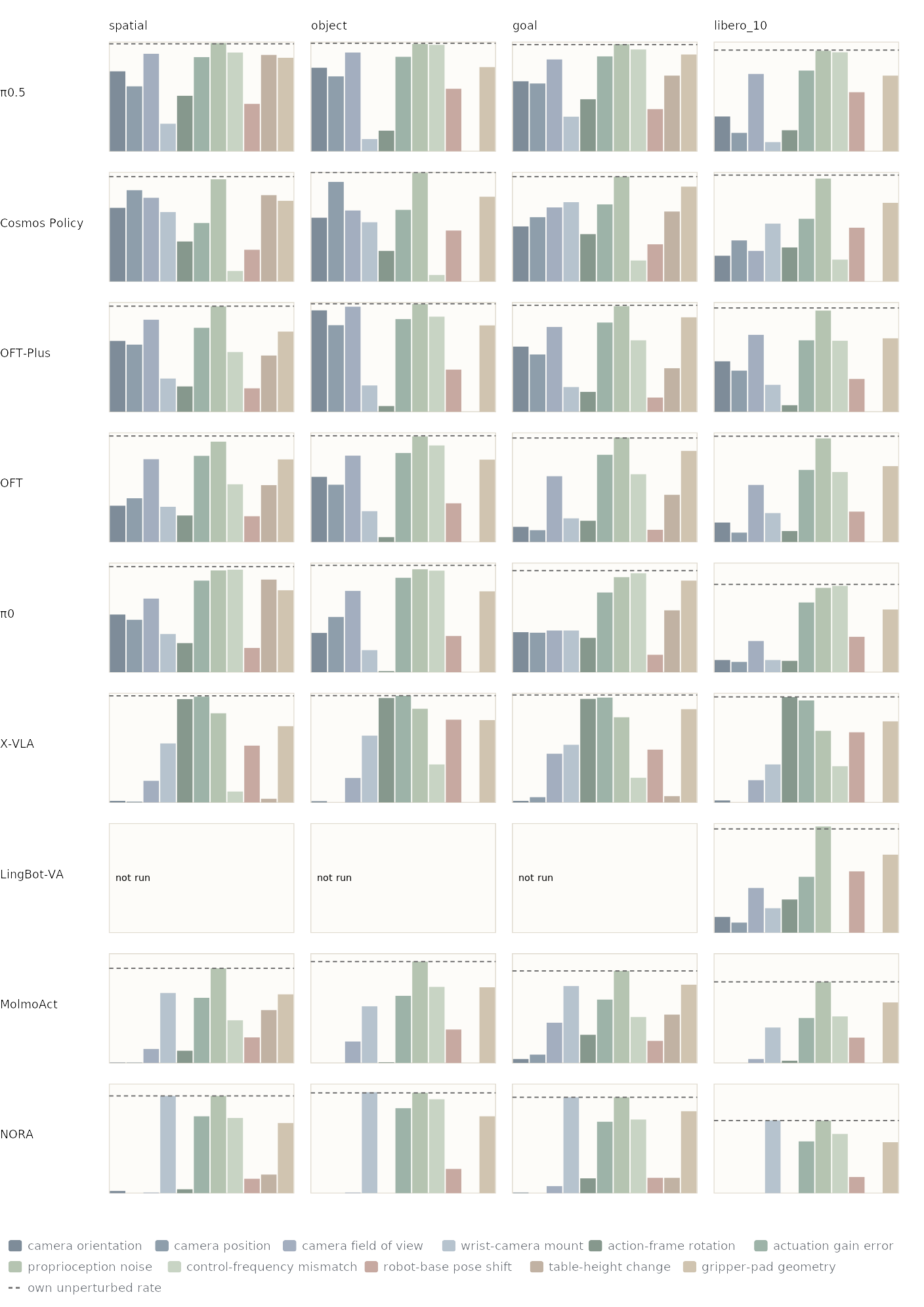}
\caption{\textbf{Full results grid, severity 5}, as in \cref{fig:grids}; the numbers are in \cref{tab:grid5}.}
\label{fig:grid_s5}
\end{figure*}

\begin{table*}[t]\centering\caption{Raw success rate per model, suite, and axis at severity 1, plotted in \cref{fig:grid_s1}. Nom.\ is the model's unperturbed rate on that suite.}\label{tab:grid1}\scriptsize\setlength{\tabcolsep}{3pt}\begin{adjustbox}{max width=\textwidth}
\begin{tabular}{@{}ll c ccccccccccc@{}}
\toprule
Model & Suite & Nom. & \rot{\axn{1} cam.\ orient.} & \rot{\axn{2} cam.\ position} & \rot{\axn{3} cam.\ FOV} & \rot{\axn{4} wrist mount} & \rot{\axn{5} frame rot.} & \rot{\axn{6} gain error} & \rot{\axn{7} ctrl.\ freq.} & \rot{\axn{8} proprio.\ noise} & \rot{\axn{9} base pose} & \rot{\axn{10} table height} & \rot{\axn{11} pad geom.} \\
\midrule
$\pi_{0.5}$ & spatial & 0.984 & 0.986 & 0.970 & 0.980 & 0.974 & 0.982 & 0.980 & 0.982 & 0.984 & 0.992 & 0.976 & 0.986 \\
 & object & 0.988 & 0.980 & 0.976 & 0.988 & 0.966 & 0.980 & 0.984 & 0.980 & 0.984 & 0.964 & n/a & 0.990 \\
 & goal & 0.976 & 0.970 & 0.972 & 0.978 & 0.978 & 0.972 & 0.974 & 0.978 & 0.976 & 0.978 & 0.968 & 0.976 \\
 & libero-10 & 0.926 & 0.922 & 0.892 & 0.918 & 0.900 & 0.922 & 0.918 & 0.928 & 0.916 & 0.922 & n/a & 0.916 \\
\addlinespace[1pt]
Cosmos Policy & spatial & 0.960 & 0.976 & 0.948 & 0.968 & 0.966 & 0.972 & 0.944 & 0.946 & 0.972 & 0.954 & 0.976 & 0.966 \\
 & object & 0.998 & 0.998 & 0.980 & 0.990 & 1.000 & 0.996 & 0.992 & 0.992 & 0.996 & 0.996 & n/a & 0.998 \\
 & goal & 0.960 & 0.974 & 0.964 & 0.976 & 0.974 & 0.968 & 0.980 & 0.954 & 0.972 & 0.970 & 0.972 & 0.956 \\
 & libero-10 & 0.974 & 0.954 & 0.954 & 0.974 & 0.964 & 0.976 & 0.968 & 0.956 & 0.978 & 0.966 & n/a & 0.960 \\
\addlinespace[1pt]
OFT-Plus & spatial & 0.966 & 0.964 & 0.952 & 0.980 & 0.954 & 0.974 & 0.962 & 0.956 & 0.978 & 0.958 & 0.948 & 0.962 \\
 & object & 0.988 & 0.986 & 0.990 & 0.980 & 0.978 & 0.978 & 0.990 & 0.988 & 0.988 & 0.958 & n/a & 0.990 \\
 & goal & 0.974 & 0.960 & 0.952 & 0.980 & 0.950 & 0.962 & 0.962 & 0.956 & 0.972 & 0.964 & 0.968 & 0.964 \\
 & libero-10 & 0.950 & 0.938 & 0.920 & 0.938 & 0.936 & 0.940 & 0.930 & 0.952 & 0.942 & 0.948 & n/a & 0.924 \\
\addlinespace[1pt]
OpenVLA-OFT & spatial & 0.970 & 0.974 & 0.942 & 0.974 & 0.956 & 0.968 & 0.974 & 0.962 & 0.982 & 0.954 & 0.972 & 0.974 \\
 & object & 0.972 & 0.966 & 0.958 & 0.978 & 0.978 & 0.978 & 0.974 & 0.980 & 0.980 & 0.952 & n/a & 0.974 \\
 & goal & 0.952 & 0.962 & 0.948 & 0.974 & 0.956 & 0.954 & 0.966 & 0.946 & 0.968 & 0.956 & 0.978 & 0.962 \\
 & libero-10 & 0.968 & 0.954 & 0.906 & 0.938 & 0.942 & 0.930 & 0.958 & 0.952 & 0.958 & 0.948 & n/a & 0.942 \\
\addlinespace[1pt]
$\pi_0$ & spatial & 0.966 & 0.956 & 0.934 & 0.978 & 0.966 & 0.942 & 0.966 & 0.962 & 0.974 & 0.960 & 0.962 & 0.956 \\
 & object & 0.978 & 0.968 & 0.934 & 0.956 & 0.966 & 0.976 & 0.980 & 0.964 & 0.974 & 0.966 & n/a & 0.980 \\
 & goal & 0.930 & 0.936 & 0.914 & 0.922 & 0.934 & 0.944 & 0.924 & 0.948 & 0.936 & 0.916 & 0.908 & 0.932 \\
 & libero-10 & 0.804 & 0.810 & 0.734 & 0.782 & 0.790 & 0.794 & 0.790 & 0.814 & 0.822 & 0.806 & n/a & 0.836 \\
\addlinespace[1pt]
X-VLA & spatial & 0.976 & 0.926 & 0.926 & 0.950 & 0.976 & 0.970 & 0.966 & 0.970 & 0.964 & 0.966 & 0.956 & 0.964 \\
 & object & 0.978 & 0.746 & 0.376 & 0.978 & 0.984 & 0.982 & 0.978 & 0.988 & 0.982 & 0.986 & n/a & 0.980 \\
 & goal & 0.984 & 0.954 & 0.982 & 0.974 & 0.976 & 0.986 & 0.980 & 0.982 & 0.984 & 0.982 & 0.960 & 0.972 \\
 & libero-10 & 0.966 & 0.790 & 0.710 & 0.934 & 0.960 & 0.952 & 0.974 & 0.944 & 0.966 & 0.972 & n/a & 0.964 \\
\addlinespace[1pt]
LingBot-VA$^\dagger$ & libero-10 & 0.950 & 0.908 & 0.862 & 0.960 & 0.962 & 0.948 & 0.952 & 0.958 & 0.966 & 0.956 & n/a & 0.962 \\
\addlinespace[1pt]
MolmoAct & spatial & 0.866 & 0.844 & 0.654 & 0.856 & 0.880 & 0.848 & 0.862 & 0.880 & 0.866 & 0.840 & 0.870 & 0.864 \\
 & object & 0.928 & 0.912 & 0.664 & 0.870 & 0.900 & 0.906 & 0.936 & 0.932 & 0.928 & 0.900 & n/a & 0.928 \\
 & goal & 0.842 & 0.872 & 0.774 & 0.892 & 0.890 & 0.848 & 0.870 & 0.850 & 0.842 & 0.838 & 0.878 & 0.824 \\
 & libero-10 & 0.742 & 0.726 & 0.548 & 0.718 & 0.776 & 0.752 & 0.766 & 0.786 & 0.742 & 0.780 & n/a & 0.740 \\
\addlinespace[1pt]
NORA & spatial & 0.890 & 0.806 & 0.572 & 0.862 & 0.890 & 0.874 & 0.884 & 0.878 & 0.890 & 0.856 & 0.876 & 0.880 \\
 & object & 0.918 & 0.532 & 0.350 & 0.856 & 0.924 & 0.918 & 0.896 & 0.932 & 0.918 & 0.906 & n/a & 0.916 \\
 & goal & 0.878 & 0.698 & 0.590 & 0.800 & 0.878 & 0.854 & 0.864 & 0.880 & 0.878 & 0.832 & 0.864 & 0.866 \\
 & libero-10 & 0.664 & 0.500 & 0.294 & 0.588 & 0.666 & 0.638 & 0.668 & 0.716 & 0.664 & 0.676 & n/a & 0.660 \\
\addlinespace[1pt]
\bottomrule
\end{tabular}
\end{adjustbox}\end{table*}
\begin{table*}[t]\centering\caption{Raw success rate per model, suite, and axis at severity 2, plotted in \cref{fig:grid_s2}.}\label{tab:grid2}\scriptsize\setlength{\tabcolsep}{3pt}\begin{adjustbox}{max width=\textwidth}
\begin{tabular}{@{}ll c ccccccccccc@{}}
\toprule
Model & Suite & Nom. & \rot{\axn{1} cam.\ orient.} & \rot{\axn{2} cam.\ position} & \rot{\axn{3} cam.\ FOV} & \rot{\axn{4} wrist mount} & \rot{\axn{5} frame rot.} & \rot{\axn{6} gain error} & \rot{\axn{7} ctrl.\ freq.} & \rot{\axn{8} proprio.\ noise} & \rot{\axn{9} base pose} & \rot{\axn{10} table height} & \rot{\axn{11} pad geom.} \\
\midrule
$\pi_{0.5}$ & spatial & 0.984 & 0.990 & 0.924 & 0.996 & 0.972 & 0.976 & 0.992 & 0.900 & 0.988 & 0.974 & 0.990 & 0.978 \\
 & object & 0.988 & 0.982 & 0.984 & 0.980 & 0.952 & 0.950 & 0.984 & 0.958 & 0.992 & 0.960 & n/a & 0.954 \\
 & goal & 0.976 & 0.974 & 0.916 & 0.976 & 0.974 & 0.954 & 0.974 & 0.940 & 0.976 & 0.918 & 0.972 & 0.970 \\
 & libero-10 & 0.926 & 0.926 & 0.818 & 0.932 & 0.866 & 0.880 & 0.912 & 0.852 & 0.918 & 0.918 & n/a & 0.866 \\
\addlinespace[1pt]
Cosmos Policy & spatial & 0.960 & 0.956 & 0.910 & 0.960 & 0.970 & 0.956 & 0.938 & 0.136 & 0.956 & 0.910 & 0.956 & 0.944 \\
 & object & 0.998 & 0.948 & 0.964 & 0.984 & 0.992 & 0.986 & 0.976 & 0.518 & 0.990 & 0.978 & n/a & 0.984 \\
 & goal & 0.960 & 0.944 & 0.912 & 0.966 & 0.966 & 0.958 & 0.964 & 0.508 & 0.960 & 0.952 & 0.974 & 0.960 \\
 & libero-10 & 0.974 & 0.914 & 0.830 & 0.936 & 0.952 & 0.936 & 0.962 & 0.436 & 0.972 & 0.950 & n/a & 0.932 \\
\addlinespace[1pt]
OFT-Plus & spatial & 0.966 & 0.976 & 0.946 & 0.976 & 0.934 & 0.958 & 0.966 & 0.524 & 0.966 & 0.860 & 0.946 & 0.952 \\
 & object & 0.988 & 0.982 & 0.994 & 0.984 & 0.960 & 0.956 & 0.984 & 0.842 & 0.986 & 0.824 & n/a & 0.976 \\
 & goal & 0.974 & 0.968 & 0.912 & 0.962 & 0.930 & 0.942 & 0.976 & 0.750 & 0.974 & 0.776 & 0.938 & 0.966 \\
 & libero-10 & 0.950 & 0.920 & 0.808 & 0.918 & 0.930 & 0.868 & 0.926 & 0.658 & 0.948 & 0.866 & n/a & 0.900 \\
\addlinespace[1pt]
OpenVLA-OFT & spatial & 0.970 & 0.930 & 0.868 & 0.972 & 0.930 & 0.950 & 0.960 & 0.602 & 0.970 & 0.830 & 0.964 & 0.956 \\
 & object & 0.972 & 0.968 & 0.940 & 0.976 & 0.964 & 0.954 & 0.972 & 0.872 & 0.974 & 0.770 & n/a & 0.960 \\
 & goal & 0.952 & 0.926 & 0.874 & 0.960 & 0.944 & 0.946 & 0.950 & 0.810 & 0.966 & 0.720 & 0.954 & 0.950 \\
 & libero-10 & 0.968 & 0.902 & 0.722 & 0.916 & 0.938 & 0.902 & 0.938 & 0.726 & 0.960 & 0.896 & n/a & 0.916 \\
\addlinespace[1pt]
$\pi_0$ & spatial & 0.966 & 0.952 & 0.852 & 0.942 & 0.958 & 0.952 & 0.954 & 0.814 & 0.970 & 0.916 & 0.968 & 0.948 \\
 & object & 0.978 & 0.960 & 0.880 & 0.944 & 0.930 & 0.918 & 0.990 & 0.918 & 0.982 & 0.914 & n/a & 0.976 \\
 & goal & 0.930 & 0.878 & 0.758 & 0.900 & 0.926 & 0.894 & 0.932 & 0.838 & 0.916 & 0.792 & 0.922 & 0.934 \\
 & libero-10 & 0.804 & 0.782 & 0.602 & 0.758 & 0.750 & 0.778 & 0.812 & 0.718 & 0.804 & 0.792 & n/a & 0.782 \\
\addlinespace[1pt]
X-VLA & spatial & 0.976 & 0.608 & 0.680 & 0.894 & 0.974 & 0.972 & 0.974 & 0.728 & 0.968 & 0.956 & 0.916 & 0.956 \\
 & object & 0.978 & 0.322 & 0.142 & 0.838 & 0.980 & 0.980 & 0.974 & 0.856 & 0.974 & 0.984 & n/a & 0.964 \\
 & goal & 0.984 & 0.692 & 0.878 & 0.960 & 0.976 & 0.974 & 0.978 & 0.856 & 0.982 & 0.978 & 0.894 & 0.970 \\
 & libero-10 & 0.966 & 0.492 & 0.358 & 0.756 & 0.958 & 0.958 & 0.962 & 0.798 & 0.964 & 0.954 & n/a & 0.914 \\
\addlinespace[1pt]
LingBot-VA$^\dagger$ & libero-10 & 0.950 & 0.766 & 0.618 & 0.906 & 0.920 & 0.924 & 0.958 & 0.184 & 0.968 & 0.952 & n/a & 0.928 \\
\addlinespace[1pt]
MolmoAct & spatial & 0.866 & 0.682 & 0.280 & 0.768 & 0.866 & 0.794 & 0.846 & 0.328 & 0.866 & 0.848 & 0.852 & 0.842 \\
 & object & 0.928 & 0.664 & 0.132 & 0.816 & 0.906 & 0.846 & 0.926 & 0.458 & 0.928 & 0.858 & n/a & 0.900 \\
 & goal & 0.842 & 0.824 & 0.534 & 0.820 & 0.852 & 0.794 & 0.838 & 0.526 & 0.842 & 0.712 & 0.858 & 0.834 \\
 & libero-10 & 0.742 & 0.586 & 0.206 & 0.606 & 0.756 & 0.660 & 0.740 & 0.294 & 0.742 & 0.700 & n/a & 0.676 \\
\addlinespace[1pt]
NORA & spatial & 0.890 & 0.578 & 0.166 & 0.640 & 0.890 & 0.840 & 0.860 & 0.566 & 0.890 & 0.788 & 0.828 & 0.870 \\
 & object & 0.918 & 0.300 & 0.056 & 0.572 & 0.924 & 0.816 & 0.890 & 0.586 & 0.918 & 0.834 & n/a & 0.920 \\
 & goal & 0.878 & 0.586 & 0.166 & 0.670 & 0.878 & 0.752 & 0.832 & 0.664 & 0.878 & 0.702 & 0.782 & 0.868 \\
 & libero-10 & 0.664 & 0.254 & 0.060 & 0.272 & 0.666 & 0.566 & 0.632 & 0.378 & 0.664 & 0.614 & n/a & 0.598 \\
\addlinespace[1pt]
\bottomrule
\end{tabular}
\end{adjustbox}\end{table*}
\begin{table*}[t]\centering\caption{Raw success rate per model, suite, and axis at severity 3, plotted in \cref{fig:grid_s3}.}\label{tab:grid3}\scriptsize\setlength{\tabcolsep}{3pt}\begin{adjustbox}{max width=\textwidth}
\begin{tabular}{@{}ll c ccccccccccc@{}}
\toprule
Model & Suite & Nom. & \rot{\axn{1} cam.\ orient.} & \rot{\axn{2} cam.\ position} & \rot{\axn{3} cam.\ FOV} & \rot{\axn{4} wrist mount} & \rot{\axn{5} frame rot.} & \rot{\axn{6} gain error} & \rot{\axn{7} ctrl.\ freq.} & \rot{\axn{8} proprio.\ noise} & \rot{\axn{9} base pose} & \rot{\axn{10} table height} & \rot{\axn{11} pad geom.} \\
\midrule
$\pi_{0.5}$ & spatial & 0.984 & 0.976 & 0.752 & 0.976 & 0.968 & 0.942 & 0.970 & 0.970 & 0.980 & 0.896 & 0.974 & 0.942 \\
 & object & 0.988 & 0.972 & 0.914 & 0.990 & 0.932 & 0.806 & 0.982 & 0.970 & 0.980 & 0.838 & n/a & 0.912 \\
 & goal & 0.976 & 0.932 & 0.734 & 0.938 & 0.954 & 0.838 & 0.966 & 0.968 & 0.984 & 0.820 & 0.964 & 0.952 \\
 & libero-10 & 0.926 & 0.854 & 0.560 & 0.914 & 0.784 & 0.762 & 0.906 & 0.910 & 0.928 & 0.862 & n/a & 0.770 \\
\addlinespace[1pt]
Cosmos Policy & spatial & 0.960 & 0.940 & 0.710 & 0.936 & 0.952 & 0.914 & 0.908 & 0.330 & 0.950 & 0.778 & 0.960 & 0.886 \\
 & object & 0.998 & 0.880 & 0.706 & 0.984 & 0.974 & 0.936 & 0.976 & 0.636 & 1.000 & 0.874 & n/a & 0.898 \\
 & goal & 0.960 & 0.858 & 0.774 & 0.944 & 0.976 & 0.894 & 0.942 & 0.594 & 0.968 & 0.758 & 0.960 & 0.944 \\
 & libero-10 & 0.974 & 0.822 & 0.466 & 0.904 & 0.924 & 0.820 & 0.902 & 0.678 & 0.970 & 0.802 & n/a & 0.818 \\
\addlinespace[1pt]
OFT-Plus & spatial & 0.966 & 0.910 & 0.874 & 0.960 & 0.868 & 0.852 & 0.952 & 0.794 & 0.978 & 0.584 & 0.896 & 0.874 \\
 & object & 0.988 & 0.990 & 0.960 & 0.982 & 0.880 & 0.786 & 0.986 & 0.930 & 0.986 & 0.606 & n/a & 0.934 \\
 & goal & 0.974 & 0.936 & 0.808 & 0.956 & 0.874 & 0.802 & 0.948 & 0.840 & 0.968 & 0.426 & 0.878 & 0.952 \\
 & libero-10 & 0.950 & 0.810 & 0.704 & 0.906 & 0.856 & 0.654 & 0.902 & 0.812 & 0.948 & 0.688 & n/a & 0.786 \\
\addlinespace[1pt]
OpenVLA-OFT & spatial & 0.970 & 0.726 & 0.578 & 0.956 & 0.846 & 0.840 & 0.938 & 0.776 & 0.972 & 0.560 & 0.938 & 0.898 \\
 & object & 0.972 & 0.916 & 0.744 & 0.968 & 0.890 & 0.784 & 0.974 & 0.936 & 0.980 & 0.610 & n/a & 0.914 \\
 & goal & 0.952 & 0.780 & 0.580 & 0.934 & 0.878 & 0.800 & 0.944 & 0.814 & 0.976 & 0.394 & 0.890 & 0.916 \\
 & libero-10 & 0.968 & 0.618 & 0.324 & 0.876 & 0.880 & 0.716 & 0.922 & 0.812 & 0.942 & 0.674 & n/a & 0.810 \\
\addlinespace[1pt]
$\pi_0$ & spatial & 0.966 & 0.880 & 0.650 & 0.910 & 0.944 & 0.866 & 0.962 & 0.964 & 0.960 & 0.648 & 0.934 & 0.878 \\
 & object & 0.978 & 0.868 & 0.600 & 0.888 & 0.888 & 0.742 & 0.964 & 0.946 & 0.972 & 0.668 & n/a & 0.902 \\
 & goal & 0.930 & 0.772 & 0.552 & 0.842 & 0.882 & 0.782 & 0.918 & 0.928 & 0.914 & 0.508 & 0.892 & 0.920 \\
 & libero-10 & 0.804 & 0.598 & 0.286 & 0.712 & 0.730 & 0.620 & 0.804 & 0.826 & 0.826 & 0.666 & n/a & 0.650 \\
\addlinespace[1pt]
X-VLA & spatial & 0.976 & 0.170 & 0.216 & 0.750 & 0.956 & 0.958 & 0.966 & 0.560 & 0.954 & 0.940 & 0.692 & 0.850 \\
 & object & 0.978 & 0.212 & 0.026 & 0.534 & 0.984 & 0.980 & 0.980 & 0.840 & 0.990 & 0.986 & n/a & 0.896 \\
 & goal & 0.984 & 0.294 & 0.392 & 0.940 & 0.962 & 0.984 & 0.972 & 0.622 & 0.970 & 0.962 & 0.588 & 0.926 \\
 & libero-10 & 0.966 & 0.126 & 0.086 & 0.668 & 0.928 & 0.962 & 0.954 & 0.760 & 0.954 & 0.856 & n/a & 0.804 \\
\addlinespace[1pt]
LingBot-VA$^\dagger$ & libero-10 & 0.950 & 0.548 & 0.364 & 0.822 & 0.872 & 0.880 & 0.914 & 0.442 & 0.966 & 0.936 & n/a & 0.802 \\
\addlinespace[1pt]
MolmoAct & spatial & 0.866 & 0.118 & 0.042 & 0.596 & 0.832 & 0.532 & 0.838 & 0.674 & 0.866 & 0.654 & 0.834 & 0.796 \\
 & object & 0.928 & 0.032 & 0.000 & 0.692 & 0.920 & 0.524 & 0.870 & 0.866 & 0.928 & 0.702 & n/a & 0.838 \\
 & goal & 0.842 & 0.496 & 0.256 & 0.726 & 0.872 & 0.634 & 0.812 & 0.632 & 0.842 & 0.536 & 0.788 & 0.802 \\
 & libero-10 & 0.742 & 0.098 & 0.000 & 0.434 & 0.780 & 0.436 & 0.686 & 0.620 & 0.742 & 0.616 & n/a & 0.610 \\
\addlinespace[1pt]
NORA & spatial & 0.890 & 0.204 & 0.010 & 0.338 & 0.890 & 0.480 & 0.860 & 0.832 & 0.890 & 0.610 & 0.744 & 0.808 \\
 & object & 0.918 & 0.034 & 0.000 & 0.254 & 0.924 & 0.446 & 0.900 & 0.916 & 0.918 & 0.644 & n/a & 0.828 \\
 & goal & 0.878 & 0.224 & 0.044 & 0.416 & 0.878 & 0.598 & 0.812 & 0.766 & 0.878 & 0.508 & 0.598 & 0.830 \\
 & libero-10 & 0.664 & 0.040 & 0.000 & 0.046 & 0.666 & 0.378 & 0.638 & 0.636 & 0.664 & 0.472 & n/a & 0.534 \\
\addlinespace[1pt]
\bottomrule
\end{tabular}
\end{adjustbox}\end{table*}
\begin{table*}[t]\centering\caption{Raw success rate per model, suite, and axis at severity 4, plotted in \cref{fig:grid_s4}.}\label{tab:grid4}\scriptsize\setlength{\tabcolsep}{3pt}\begin{adjustbox}{max width=\textwidth}
\begin{tabular}{@{}ll c ccccccccccc@{}}
\toprule
Model & Suite & Nom. & \rot{\axn{1} cam.\ orient.} & \rot{\axn{2} cam.\ position} & \rot{\axn{3} cam.\ FOV} & \rot{\axn{4} wrist mount} & \rot{\axn{5} frame rot.} & \rot{\axn{6} gain error} & \rot{\axn{7} ctrl.\ freq.} & \rot{\axn{8} proprio.\ noise} & \rot{\axn{9} base pose} & \rot{\axn{10} table height} & \rot{\axn{11} pad geom.} \\
\midrule
$\pi_{0.5}$ & spatial & 0.984 & 0.912 & 0.722 & 0.960 & 0.440 & 0.866 & 0.960 & 0.316 & 0.984 & 0.582 & 0.966 & 0.888 \\
 & object & 0.988 & 0.964 & 0.666 & 0.956 & 0.308 & 0.550 & 0.974 & 0.452 & 0.990 & 0.708 & n/a & 0.832 \\
 & goal & 0.976 & 0.900 & 0.608 & 0.920 & 0.528 & 0.706 & 0.952 & 0.638 & 0.982 & 0.670 & 0.834 & 0.924 \\
 & libero-10 & 0.926 & 0.650 & 0.354 & 0.836 & 0.218 & 0.570 & 0.876 & 0.458 & 0.924 & 0.718 & n/a & 0.722 \\
\addlinespace[1pt]
Cosmos Policy & spatial & 0.960 & 0.852 & 0.768 & 0.856 & 0.896 & 0.750 & 0.876 & 0.002 & 0.962 & 0.458 & 0.936 & 0.780 \\
 & object & 0.998 & 0.710 & 0.832 & 0.930 & 0.936 & 0.736 & 0.960 & 0.000 & 0.996 & 0.582 & n/a & 0.856 \\
 & goal & 0.960 & 0.698 & 0.682 & 0.852 & 0.948 & 0.808 & 0.886 & 0.148 & 0.970 & 0.586 & 0.868 & 0.902 \\
 & libero-10 & 0.974 & 0.598 & 0.378 & 0.702 & 0.856 & 0.616 & 0.866 & 0.008 & 0.966 & 0.660 & n/a & 0.708 \\
\addlinespace[1pt]
OFT-Plus & spatial & 0.966 & 0.812 & 0.684 & 0.922 & 0.636 & 0.538 & 0.908 & 0.010 & 0.970 & 0.336 & 0.756 & 0.774 \\
 & object & 0.988 & 0.974 & 0.872 & 0.980 & 0.600 & 0.480 & 0.966 & 0.006 & 0.988 & 0.542 & n/a & 0.864 \\
 & goal & 0.974 & 0.876 & 0.594 & 0.902 & 0.592 & 0.590 & 0.938 & 0.236 & 0.976 & 0.314 & 0.604 & 0.916 \\
 & libero-10 & 0.950 & 0.700 & 0.474 & 0.828 & 0.564 & 0.454 & 0.846 & 0.088 & 0.946 & 0.480 & n/a & 0.694 \\
\addlinespace[1pt]
OpenVLA-OFT & spatial & 0.970 & 0.580 & 0.434 & 0.856 & 0.642 & 0.574 & 0.938 & 0.022 & 0.960 & 0.342 & 0.754 & 0.790 \\
 & object & 0.972 & 0.694 & 0.504 & 0.942 & 0.624 & 0.522 & 0.952 & 0.012 & 0.986 & 0.468 & n/a & 0.864 \\
 & goal & 0.952 & 0.508 & 0.232 & 0.820 & 0.536 & 0.586 & 0.896 & 0.224 & 0.976 & 0.272 & 0.632 & 0.894 \\
 & libero-10 & 0.968 & 0.364 & 0.122 & 0.768 & 0.614 & 0.480 & 0.880 & 0.102 & 0.948 & 0.460 & n/a & 0.684 \\
\addlinespace[1pt]
$\pi_0$ & spatial & 0.966 & 0.702 & 0.438 & 0.822 & 0.666 & 0.672 & 0.954 & 0.214 & 0.946 & 0.340 & 0.946 & 0.792 \\
 & object & 0.978 & 0.680 & 0.430 & 0.812 & 0.526 & 0.384 & 0.956 & 0.280 & 0.978 & 0.528 & n/a & 0.824 \\
 & goal & 0.930 & 0.584 & 0.370 & 0.690 & 0.686 & 0.644 & 0.904 & 0.468 & 0.940 & 0.396 & 0.694 & 0.872 \\
 & libero-10 & 0.804 & 0.320 & 0.114 & 0.522 & 0.308 & 0.434 & 0.744 & 0.328 & 0.814 & 0.470 & n/a & 0.576 \\
\addlinespace[1pt]
X-VLA & spatial & 0.976 & 0.036 & 0.010 & 0.550 & 0.690 & 0.964 & 0.978 & 0.000 & 0.964 & 0.664 & 0.260 & 0.774 \\
 & object & 0.978 & 0.082 & 0.002 & 0.246 & 0.832 & 0.980 & 0.972 & 0.000 & 0.962 & 0.902 & n/a & 0.818 \\
 & goal & 0.984 & 0.038 & 0.046 & 0.690 & 0.694 & 0.976 & 0.966 & 0.224 & 0.926 & 0.892 & 0.192 & 0.876 \\
 & libero-10 & 0.966 & 0.026 & 0.010 & 0.422 & 0.592 & 0.958 & 0.954 & 0.000 & 0.922 & 0.788 & n/a & 0.720 \\
\addlinespace[1pt]
LingBot-VA$^\dagger$ & libero-10 & 0.950 & 0.398 & 0.102 & 0.672 & 0.568 & 0.766 & 0.836 & 0.000 & 0.964 & 0.806 & n/a & 0.714 \\
\addlinespace[1pt]
MolmoAct & spatial & 0.866 & 0.016 & 0.010 & 0.330 & 0.712 & 0.292 & 0.788 & 0.010 & 0.866 & 0.366 & 0.698 & 0.710 \\
 & object & 0.928 & 0.000 & 0.000 & 0.362 & 0.812 & 0.162 & 0.822 & 0.006 & 0.928 & 0.500 & n/a & 0.762 \\
 & goal & 0.842 & 0.160 & 0.118 & 0.504 & 0.772 & 0.494 & 0.752 & 0.180 & 0.842 & 0.372 & 0.582 & 0.770 \\
 & libero-10 & 0.742 & 0.000 & 0.000 & 0.184 & 0.454 & 0.228 & 0.620 & 0.018 & 0.742 & 0.392 & n/a & 0.526 \\
\addlinespace[1pt]
NORA & spatial & 0.890 & 0.074 & 0.000 & 0.088 & 0.890 & 0.204 & 0.834 & 0.112 & 0.890 & 0.286 & 0.464 & 0.704 \\
 & object & 0.918 & 0.000 & 0.000 & 0.038 & 0.924 & 0.082 & 0.852 & 0.068 & 0.918 & 0.370 & n/a & 0.772 \\
 & goal & 0.878 & 0.130 & 0.012 & 0.162 & 0.878 & 0.414 & 0.784 & 0.240 & 0.878 & 0.312 & 0.310 & 0.790 \\
 & libero-10 & 0.664 & 0.002 & 0.000 & 0.002 & 0.666 & 0.166 & 0.604 & 0.040 & 0.664 & 0.290 & n/a & 0.502 \\
\addlinespace[1pt]
\bottomrule
\end{tabular}
\end{adjustbox}\end{table*}
\begin{table*}[t]\centering\caption{Raw success rate per model, suite, and axis at severity 5, plotted in \cref{fig:grid_s5}.}\label{tab:grid5}\scriptsize\setlength{\tabcolsep}{3pt}\begin{adjustbox}{max width=\textwidth}
\begin{tabular}{@{}ll c ccccccccccc@{}}
\toprule
Model & Suite & Nom. & \rot{\axn{1} cam.\ orient.} & \rot{\axn{2} cam.\ position} & \rot{\axn{3} cam.\ FOV} & \rot{\axn{4} wrist mount} & \rot{\axn{5} frame rot.} & \rot{\axn{6} gain error} & \rot{\axn{7} ctrl.\ freq.} & \rot{\axn{8} proprio.\ noise} & \rot{\axn{9} base pose} & \rot{\axn{10} table height} & \rot{\axn{11} pad geom.} \\
\midrule
$\pi_{0.5}$ & spatial & 0.984 & 0.732 & 0.594 & 0.892 & 0.252 & 0.508 & 0.862 & 0.904 & 0.990 & 0.434 & 0.882 & 0.856 \\
 & object & 0.988 & 0.764 & 0.686 & 0.904 & 0.112 & 0.188 & 0.864 & 0.974 & 0.982 & 0.572 & n/a & 0.770 \\
 & goal & 0.976 & 0.640 & 0.620 & 0.840 & 0.316 & 0.476 & 0.868 & 0.932 & 0.980 & 0.386 & 0.692 & 0.886 \\
 & libero-10 & 0.926 & 0.318 & 0.168 & 0.708 & 0.084 & 0.192 & 0.738 & 0.906 & 0.920 & 0.540 & n/a & 0.692 \\
\addlinespace[1pt]
Cosmos Policy & spatial & 0.960 & 0.674 & 0.836 & 0.766 & 0.636 & 0.366 & 0.536 & 0.096 & 0.936 & 0.290 & 0.790 & 0.738 \\
 & object & 0.998 & 0.584 & 0.912 & 0.650 & 0.542 & 0.280 & 0.656 & 0.060 & 0.994 & 0.466 & n/a & 0.776 \\
 & goal & 0.960 & 0.504 & 0.588 & 0.678 & 0.726 & 0.434 & 0.706 & 0.192 & 0.960 & 0.340 & 0.640 & 0.868 \\
 & libero-10 & 0.974 & 0.236 & 0.376 & 0.280 & 0.530 & 0.312 & 0.574 & 0.200 & 0.942 & 0.492 & n/a & 0.720 \\
\addlinespace[1pt]
OFT-Plus & spatial & 0.966 & 0.648 & 0.614 & 0.842 & 0.304 & 0.232 & 0.768 & 0.546 & 0.964 & 0.214 & 0.514 & 0.734 \\
 & object & 0.988 & 0.928 & 0.792 & 0.962 & 0.240 & 0.052 & 0.848 & 0.870 & 0.986 & 0.386 & n/a & 0.790 \\
 & goal & 0.974 & 0.596 & 0.524 & 0.776 & 0.226 & 0.182 & 0.816 & 0.654 & 0.966 & 0.130 & 0.398 & 0.864 \\
 & libero-10 & 0.950 & 0.462 & 0.376 & 0.704 & 0.246 & 0.060 & 0.654 & 0.650 & 0.926 & 0.300 & n/a & 0.672 \\
\addlinespace[1pt]
OpenVLA-OFT & spatial & 0.970 & 0.332 & 0.400 & 0.758 & 0.322 & 0.242 & 0.788 & 0.528 & 0.918 & 0.236 & 0.520 & 0.756 \\
 & object & 0.972 & 0.596 & 0.524 & 0.790 & 0.282 & 0.044 & 0.814 & 0.884 & 0.966 & 0.354 & n/a & 0.754 \\
 & goal & 0.952 & 0.138 & 0.108 & 0.602 & 0.216 & 0.194 & 0.798 & 0.620 & 0.956 & 0.112 & 0.432 & 0.834 \\
 & libero-10 & 0.968 & 0.178 & 0.086 & 0.522 & 0.264 & 0.100 & 0.660 & 0.640 & 0.948 & 0.278 & n/a & 0.694 \\
\addlinespace[1pt]
$\pi_0$ & spatial & 0.966 & 0.528 & 0.480 & 0.674 & 0.350 & 0.266 & 0.838 & 0.938 & 0.932 & 0.222 & 0.848 & 0.750 \\
 & object & 0.978 & 0.360 & 0.506 & 0.744 & 0.202 & 0.010 & 0.864 & 0.930 & 0.942 & 0.332 & n/a & 0.740 \\
 & goal & 0.930 & 0.366 & 0.362 & 0.382 & 0.382 & 0.314 & 0.730 & 0.906 & 0.870 & 0.160 & 0.566 & 0.838 \\
 & libero-10 & 0.804 & 0.112 & 0.094 & 0.286 & 0.112 & 0.104 & 0.638 & 0.790 & 0.772 & 0.324 & n/a & 0.574 \\
\addlinespace[1pt]
X-VLA & spatial & 0.976 & 0.014 & 0.008 & 0.198 & 0.540 & 0.946 & 0.968 & 0.100 & 0.816 & 0.520 & 0.034 & 0.698 \\
 & object & 0.978 & 0.012 & 0.000 & 0.224 & 0.612 & 0.956 & 0.976 & 0.348 & 0.858 & 0.758 & n/a & 0.754 \\
 & goal & 0.984 & 0.014 & 0.048 & 0.446 & 0.528 & 0.948 & 0.960 & 0.226 & 0.780 & 0.484 & 0.058 & 0.854 \\
 & libero-10 & 0.966 & 0.018 & 0.000 & 0.204 & 0.348 & 0.964 & 0.934 & 0.332 & 0.656 & 0.642 & n/a & 0.742 \\
\addlinespace[1pt]
LingBot-VA$^\dagger$ & libero-10 & 0.950 & 0.144 & 0.092 & 0.410 & 0.224 & 0.304 & 0.512 & 0.004 & 0.972 & 0.562 & n/a & 0.714 \\
\addlinespace[1pt]
MolmoAct & spatial & 0.866 & 0.004 & 0.004 & 0.128 & 0.640 & 0.112 & 0.596 & 0.390 & 0.866 & 0.234 & 0.484 & 0.628 \\
 & object & 0.928 & 0.000 & 0.000 & 0.196 & 0.518 & 0.006 & 0.614 & 0.696 & 0.928 & 0.306 & n/a & 0.692 \\
 & goal & 0.842 & 0.036 & 0.076 & 0.368 & 0.704 & 0.258 & 0.580 & 0.420 & 0.842 & 0.202 & 0.442 & 0.716 \\
 & libero-10 & 0.742 & 0.000 & 0.000 & 0.036 & 0.324 & 0.020 & 0.412 & 0.426 & 0.742 & 0.232 & n/a & 0.554 \\
\addlinespace[1pt]
NORA & spatial & 0.890 & 0.020 & 0.000 & 0.006 & 0.890 & 0.036 & 0.704 & 0.688 & 0.890 & 0.132 & 0.170 & 0.642 \\
 & object & 0.918 & 0.000 & 0.000 & 0.006 & 0.924 & 0.000 & 0.778 & 0.860 & 0.918 & 0.222 & n/a & 0.704 \\
 & goal & 0.878 & 0.006 & 0.000 & 0.064 & 0.878 & 0.136 & 0.654 & 0.674 & 0.878 & 0.140 & 0.140 & 0.750 \\
 & libero-10 & 0.664 & 0.000 & 0.000 & 0.000 & 0.666 & 0.000 & 0.474 & 0.542 & 0.664 & 0.148 & n/a & 0.466 \\
\addlinespace[1pt]
\bottomrule
\end{tabular}
\end{adjustbox}\end{table*}

\section{Additional Findings from the Profiles}
\label{sec:app_more}
\textbf{Control-frequency mismatch damages policies by overshoot.} A 20 Hz control period realizes about 17 percent of a requested displacement, so policies learn to ask for near the 5 cm cap; a 5 Hz period realizes about 64 percent, the policy cannot tell, and the error compounds across the action chunk, so a quarter of the trained rate costs 0.78 of pooled success. For OpenVLA-OFT, four times the simulated time recovers little (0.099 to 0.206 of its nominal) while a fixed interval between observations recovers most (0.803); \cref{sec:supp_a4} gives the decomposition. The two models that predict video jointly with their actions, Cosmos Policy and LingBot-VA, sit at the bottom of this axis (0.419, the worst four-suite rate, and 0.318 on LIBERO-10), which suggests that training on video at a fixed frame rate ties a policy's actions to that clock, a hypothesis resting on two rows.

\textbf{The action interface is associated with sensitivity to different faults.} X-VLA outputs absolute poses, and the calibration faults that hurt every other model have little effect on it, 0.968 on action-frame rotation and on gain error against a 0.976 nominal, since an absolute goal shrinks at every step the gap these faults distort, whereas a delta policy's errors accumulate. The same interface offers no apparent protection against perturbations to the scene cameras (0.329 and 0.295) and pays on control frequency (0.557), so X-VLA lands sixth of eight. Training data and chunk length differ too, so this is an association rather than a controlled comparison.

\textbf{Robustness training transfers only to the axes represented in training.} OFT-Plus beats OpenVLA-OFT overall, 0.796 against 0.761 from near-identical nominals, and the gains are concentrated on the camera axes its training mixture perturbs: camera orientation 0.867 against 0.701, camera position 0.787 against 0.590, field of view 0.922 against 0.874. On every other axis, including the wrist camera, which the mixture leaves unperturbed, the two differ by at most 0.02; a benchmark of camera perturbations alone would call OFT-Plus robust.

\section{Statistical Methods}
\label{sec:supp_stats}
Two properties of the data drive every choice below. The data are clustered: a condition holds 500 episodes from 10 tasks, and the episodes within a task have correlated outcomes, so the unit that varies independently is the task. On the same statistic, the pooled severity-3 rate over 168,000 episodes, treating episodes as independent gives a 95 percent interval of 0.765 to 0.769; with the task as the unit it is \ciLoc{} to \ciHic, about twelve times wider. The data are paired: every model is evaluated on the same 40 tasks, from the same initial states, under perturbations pinned in the frozen manifests, so task-to-task difficulty cancels when models or axes are compared within a block. Everything below follows from those two facts. One more convention: pooled rates weight episodes equally, so a suite with a table-height axis (spatial, goal) carries 11 of 42 conditions at a level and the other two suites 10 of 42; weighting the four suites equally instead changes every pooled level by at most 0.001, and the leaderboard weights axes, not episodes, equally. Unless otherwise specified, analyses use 10,000 permutations, 5,000 bootstrap draws, and a fixed seed, and every input is a per-condition result file holding successes and trials per task, so every number can be recomputed from episode counts.

\paragraph{Cluster bootstrap over tasks~\cite{efron1993bootstrap,cameron2015practitioner}.}
Every confidence interval resamples the 10 tasks of a suite with replacement, all 50 episodes of a chosen task retained together, and recomputes the statistic; the middle 95 percent of 5,000 recomputations is the interval. Within a suite, one draw is shared across every model, axis, and severity, which is what makes comparisons paired; draws are independent between suites, because task $i$ of one suite is unrelated to task $i$ of another. Drawing independently per model injects variation that does not exist in the experiment: on the goal suite it gave a rank-1 frequency of 0.83 for \pihalf against 0.99 with shared draws. Caveat: ten clusters are few. With ten task clusters per suite, the finite-sample coverage of these percentile intervals is not established; the few-clusters literature warns that coverage can degrade and gives no universal threshold~\cite{cameron2015practitioner}; we report the percentile intervals without a simulation-based coverage check or a wild cluster bootstrap~\cite{cameron2008wild}.

\paragraph{Permutation test over blocks~\cite{good2005permutation}.}
Do the 11 axes hurt equally? Every (model, suite) pair is a block holding one success rate per axis available in that suite, all measured on the same tasks: 32 blocks, with table height present in the spatial and goal blocks only. Under the null that axis identity carries no information, the labels inside a block are exchangeable, so we shuffle them within each block, recompute the axis means over the blocks that carry each axis, and record their variance; $p = (k+1)/(B+1)$, where $k$ counts shuffles at least as spread out as the observation. No shuffle reached the observed spread, so $p = 1/10{,}001$, which we report as $p<0.001$. A repeated-measures ANOVA would assume normal residuals and equal variances; these are bounded proportions, many at exactly 0 or 1. The test conditions on the design: blocks share tasks and include related models, and the frozen manifests ensure that every axis was measured on the same tasks and starts, which is what the within-block shuffle relies on. We read it as descriptive of this roster.

\paragraph{Pairwise sign-flip tests, Holm correction, and letters~\cite{holm1979,piepho2004}.}
For each of the 55 axis pairs, take the per-block difference and compare its mean with the distribution under random sign flips: exact enumeration of all $2^{16}$ patterns for pairs involving table height (16 blocks), 200,000 random flips otherwise, with $p = (k+1)/(B+1)$. Holm's step-down corrects the 55 tests, and \sigPairs{} pairs are separated. The compact letter display (\cref{tab:letters}) is built from the significance matrix as the maximal sets of mutually unseparated axes, so two axes share a letter exactly when the tests do not separate them~\cite{piepho2004}; \letterGroups{} letters are needed. The letters are not a restatement of the ranking. Camera orientation and control-frequency mismatch differ by 0.190 in their means and are not separated (Holm-adjusted $p = 0.078$): the per-block difference has a standard deviation of 0.346, and only 23 of the 32 blocks agree on which axis is worse, because at severity 3 the control-frequency rung is 50 Hz, where Cosmos Policy collapses and the other models are barely affected. Actuation gain error and proprioception noise differ by only 0.023 and are separated: the per-block difference has a standard deviation of 0.021 and 25 of 32 blocks agree in sign. In a paired design, significance depends on both the magnitude and variability of the paired differences. The same test, over the 10 task differences within a suite with exact enumeration of the 1,024 sign patterns and Holm correction, compares the top model with each rival in \cref{tab:rq3}; on the short-horizon suites the only rival it does not separate from \pihalf is Cosmos Policy on goal (Holm-adjusted $p = 0.062$). Because the letters overlap, they are groups rather than tiers, and their number changes with the level (\letterGroupsSfour{} at level 4, \letterGroupsSfive{} at level 5).

\begin{table}[t]
\centering
\caption{\textbf{Axis ordering at severity 3 with compact letters.} Raw success pooled over the eight four-suite models. Letters are the maximal sets of axes that the Holm-corrected paired sign-flip tests do not separate, so two axes share a letter exactly when they are not separated; \sigPairs{} of the 55 pairs are separated. The groups overlap, so they are not tiers.}
\label{tab:letters}
\small
\begin{tabular}{@{}lcc@{}}
\toprule
Axis & Success at s3 & Letters \\
\midrule
camera position & 0.448 & a \\
camera orientation & 0.596 & b \\
robot-base pose shift & 0.692 & bc \\
action-frame rotation & 0.752 & bc \\
camera field of view & 0.777 & cde \\
control-frequency mismatch & 0.786 & bce \\
table-height change & 0.846 & cdef \\
gripper-pad geometry & 0.846 & e \\
wrist-camera mount & 0.890 & df \\
actuation gain error & 0.908 & f \\
proprioception noise & 0.931 & g \\
\bottomrule
\end{tabular}

\end{table}

\paragraph{Non-inferiority~\cite{piantadosi2017clinical,altman1995absence}.}
``This severity level is harmless'' is a positive claim and needs positive evidence; a failed significance test cannot distinguish a harmless level from an underpowered experiment. We choose a margin of 5 percent of unperturbed performance, bootstrap the ratio of the level's pooled rate to the nominal rate with the task-clustered method, and require the entire 95 percent interval to lie above 0.95. Severity 1 passes: ratio \ratioa{} with interval [\ratioLoa, \ratioHia]. Severity 2 fails: ratio \ratiob{} with interval [\ratioLob, \ratioHib], which also lies below a 10 percent margin, so the verdict does not hinge on the choice between 5 and 10 percent. The margin is a convention rather than a fact, and passing it bounds the loss rather than proving its absence.

\paragraph{Severity trend within blocks~\cite{hollander2013nonparametric}.}
Does damage rise with severity inside each block, or only in the average? Each (model, suite, axis) block holds one rate per level; there are \trendBlocks{} such blocks with control frequency excluded, since its ladder is not ordered by dose. Within each block we compute the Spearman correlation between level and success with average ranks for ties; \trendConst{} blocks are constant across all five levels (NORA on the wrist-camera and proprioception axes, MolmoAct on proprioception, inputs that never reach the network) and have no defined correlation. Of the remaining 292, \trendNeg{} are negative and 5 positive; the mean correlation is \trendMeanRho, and none of 10,000 within-block permutations of the levels reaches it ($p = 1/10{,}001$). The descriptive counts are weaker than a monotone decline: \trendNonInc{} of the 304 blocks are non-increasing across all five levels (including the constant ones) and \trendStrict{} decrease strictly; the rest fall overall with a local reversal, for example \pizero{} on spatial camera position (0.934, 0.852, 0.650, 0.438, 0.480). None of the \trendExceptions{} blocks without a negative correlation shows a harsher rig helping: the largest severity-5 minus severity-1 gap among them is $+0.012$, on conditions of 500 episodes where one standard error is about 0.01. Levels are ordered but not on a common scale across axes, which is why the analysis uses ranks and nothing more.

\paragraph{Exact Wilcoxon signed-rank test~\cite{wilcoxon1945}.}
Is LIBERO-10 harder for every model? Each model contributes one number, its LIBERO-10 gap over the ten shared axes, and the exact Wilcoxon signed-rank test enumerates all $2^8 = 256$ sign assignments. All eight gaps are negative, the most extreme outcome available, so $p = 0.008$ at every level checked. The eight models are a chosen roster that includes two OpenVLA-OFT variants, not a sample from a population, so the test describes this roster; the gap also mixes task length with lower nominal rates and different task content, which the raw-rate metric does not separate.

\section{Control-Frequency Decomposition}
\label{sec:supp_a4}
Changing the control rate under a fixed simulated-time budget changes two things at once: how many decisions the policy gets, and how long it executes without new observations between decisions. Two diagnostic executor rules each freeze exactly one, on the same 20 conditions, the same frozen manifests, and the same paired starting states, at the screen tier (\cref{tab:a4}). V1 asks whether the problem is too little time; V2 asks whether it is too long spent blind. V1 is a sensitive instrument when a real time limit exists: freezing the step count at 100 Hz leaves 2.2 seconds of simulated time, and there OpenVLA-OFT falls from 0.707 to 0.034. At 5 Hz, where V1 hands the policy four times the simulated time, it recovers only 0.107, so the budget matters little for OpenVLA-OFT at that rate. V2's re-query interval is clamped to the chunk length, so the actual number of actions executed per query was 2, 4, 8, 8, 8 for OpenVLA-OFT at 5, 10, 25, 50, and 100 Hz, and 1, 2, 6, 10, 10 for \pihalf; at 50 and 100 Hz the clamp makes V2 identical to the official OpenVLA-OFT run, and those conditions reproduce it to three decimals, which checks that V2 changes nothing else. The split is model dependent: at 5 Hz the time limit accounts for about 12 percent of OpenVLA-OFT's deficit and blind execution for about 78 percent, while for \pihalf the time limit accounts for about 61 percent, because \pihalf natively re-queries every 5 steps and so was never as blind. These shares come from two single interventions that interact, so they are not an additive decomposition. What holds for both is the ranking: looking more often helps more than having more time. The high-rate side carries its own finding: \pihalf scores 0.929 at 100 Hz officially, and forcing it to execute full chunks drops it to 0.490, so its strength at high rates depends on re-planning often, not by tolerating the rate.

\begin{table}[t]
\centering
\caption{\textbf{Control-frequency decomposition.} Success relative to the model's own nominal, screen tier (10 trials per task) on the same 20 conditions and manifests. V0 is the official protocol: the step budget scales with the rate so wall-clock time is constant, and the policy executes its native chunk. V1 freezes the step count at the 20 Hz cap, so the simulated time floats (4$\times$ at 5 Hz, 0.2$\times$ at 100 Hz). V2 keeps the official budget but re-queries the policy at a fixed simulated-time interval (0.4 s for OpenVLA-OFT, 0.25 s for \pihalf), clamped to the chunk length: 2, 4, 8, 8, 8 and 1, 2, 6, 10, 10 actions per query at 5, 10, 25, 50, and 100 Hz.}
\label{tab:a4}
\small
\begin{adjustbox}{max width=\linewidth}\begin{tabular}{@{}lccc@{}}
\toprule
OpenVLA-OFT, per rate & V0 official & V1 step budget & V2 replan in time \\
\midrule
5 Hz (level 4) & 0.099 & 0.206 & \textbf{0.803} \\
10 Hz (level 2) & 0.736 & 0.795 & 0.930 \\
25 Hz (level 1) & 1.002 & 0.997 & 1.002 \\
50 Hz (level 3) & 0.888 & 0.505 & 0.888 \\
100 Hz (level 5) & 0.707 & \textbf{0.034} & 0.707 \\
axis mean & 0.686 & 0.507 & 0.866 \\
\midrule
\pihalf, per rate & & & \\
\midrule
5 Hz (level 4) & 0.500 & 0.804 & \textbf{0.890} \\
10 Hz (level 2) & 0.936 & 0.982 & 0.949 \\
25 Hz (level 1) & 0.982 & 0.992 & 0.997 \\
50 Hz (level 3) & 0.964 & 0.489 & 0.804 \\
100 Hz (level 5) & 0.929 & \textbf{0.036} & 0.490 \\
axis mean & 0.862 & 0.661 & 0.826 \\
\bottomrule
\end{tabular}\end{adjustbox}
\end{table}

\paragraph{The overshoot probe and its limits.}
The figures 17 percent, 64 percent, 4.8 cm, 17.4 cm, and 3.6$\times$ come from replaying one real eight-action chunk (OpenVLA-OFT, confirm tier, spatial task ``pick up the black bowl between the plate and the ramekin and place it on the plate'', episode 0) from one initial state at both rates and logging the end-effector displacement per step: net travel 4.77 cm at 20 Hz and 17.36 cm at 5 Hz. It is one chunk of one episode of one task, so it illustrates the mechanism and does not estimate a population. The trajectory-deviation figure is the least stable of the set (12.95 and 12.12 cm on two identical invocations, with a 1.4 to 2.2 cm head start because the three settling steps take 150 ms at 20 Hz and 600 ms at 5 Hz), which is why the text says roughly 13 cm. Speed is not what is lost: each command travels 3.7 times farther but commands arrive a quarter as often, so achievable speed at 5 Hz is about 94 percent of nominal.

\paragraph{Two clocks.}
One action is consumed per control step, 50 ms at 20 Hz and 200 ms at 5 Hz, and that is when the delta re-anchors on the arm's true measured pose. The network runs far less often: once per chunk, so every 8 control steps for OpenVLA-OFT, which is 400 ms at 20 Hz and 1.6 s at 5 Hz. Re-anchoring is on the fast clock, but new visual information arrives on the slow clock, so every step executes a command based on an outdated observation. This is also why V2 works without touching the servo: it shortens only the slow clock.

\paragraph{Retired explanations.}
Two explanations were ruled out by these runs. ``The task stops fitting in the speed budget'' fails because the realization fraction is not constant with rate, so achievable speed is roughly preserved. ``The world moves on during the long chunk'' fails because LIBERO is quasi-static: objects do not evolve on their own, so there is no physical deadline for an action to miss. A secondary channel is visible but not separately quantified: closing the gripper takes about 6 steps at 20 Hz and completes inside 1 step at 5 Hz, which compresses the learned grasp-phase coordination.

\section{Proprioception: Per-Model Reading and Follow-Ups}
\label{sec:supp_proprio}
\cref{tab:proprio} lists each model's unperturbed rate and its rate at the harshest proprioception dose (5 cm of position noise and 8 degrees of orientation noise). Three groups appear. Exact equality with nominal (NORA, MolmoAct) is not robustness: their inference code never passes the state into the network, so there is nothing for the perturbation to reach. A small gap (\pihalf through \pizero) means the state reaches the network but changes its output little: at step 0 the camera images are identical to the unperturbed run, so any change in the first action comes from the corrupted numbers alone, and \pihalf's first action moves by 0.019 (the norm of the seven-dimensional action change), equal to its own sampling noise floor, while OpenVLA-OFT's moves by 0.006 at severity 3 and 0.023 at severity 5, about 2 and 7 percent of a typical action norm of 0.33. A large gap (X-VLA alone, $-0.199$) means the state is genuinely used: X-VLA commands absolute hand positions and must anchor its goals on the sensed state, and its first action changes by 0.436 under the harshest corruption. The follow-ups reported in the main text (random and frozen state reports; noise injected into the controller's belief) were run on OpenVLA-OFT on the spatial suite, 100 episodes per dose; the report-only rates at the same doses were 0.982, 0.970, 0.972, and 0.960 against the in-loop 0.990, 0.980, 0.890, and 0.470. For this policy and suite, in-loop noise below 5 mm caused no observed degradation and 2 cm halved success; a general noise budget would require evaluation across more models and suites. Two rows serve as no-op controls for the harness. NORA does not read the wrist camera: its five wrist-mount conditions have identical episode outcomes on every suite, and its proprioception conditions match its nominal episode for episode, yet the wrist-mount outcomes differ from nominal in 30 of 2,000 episodes (aggregate differences of 0.000, 0.006, 0.000, and 0.002 across the four suites, a net gain of four successes). LingBot-VA does not read the state, and its harshest proprioception condition is 0.022 above its nominal on LIBERO-10. We have not isolated the cause of these no-op discrepancies, which could be a systematic difference in the perturbed code path or numerical nondeterminism, so we do not interpret them as policy responses, and no difference of that size is interpreted in this paper.

\begin{table}[t]
\centering
\caption{\textbf{What each model does with a corrupted position report.} Confirm tier. $^\dagger$single suite.}
\label{tab:proprio}
\small
\begin{tabular}{@{}lccc@{}}
\toprule
Model & Nominal & Harshest dose (s5) & Change \\
\midrule
$\pi_{0.5}$ & 0.969 & 0.968 & $-0.001$ \\
Cosmos Policy & 0.973 & 0.958 & $-0.015$ \\
OFT-Plus & 0.970 & 0.961 & $-0.009$ \\
OpenVLA-OFT & 0.966 & 0.947 & $-0.019$ \\
$\pi_0$ & 0.919 & 0.879 & $-0.040$ \\
X-VLA & 0.976 & 0.777 & $-0.199$ \\
LingBot-VA$^\dagger$ & 0.950 & 0.972 & $+0.022$ \\
MolmoAct & 0.845 & 0.845 & $+0.000$ \\
NORA & 0.838 & 0.838 & $+0.000$ \\
\bottomrule
\end{tabular}

\end{table}

\section{Fairness Checks and the Scoring Decision}
\label{sec:supp_diag}
\paragraph{Feasibility witnesses.}
Compensated replay (action-frame rotation at levels 3 and 5, gain error at level 5, on every suite) reaches 93 to 107 percent of the nominal replay rate (baseline replay rates: spatial 0.85, object 0.90, goal 0.85, LIBERO-10 0.70); each rate is estimated from 20 replays per suite (two demonstrations per task), so one replay moves a rate by 5 percentage points and values above 100 percent are sampling noise, not a benefit of the fault. The closed-loop tracking witness (control frequency, base pose, table height) is certified when its rate reaches 0.8 times the suite's nominal replay rate for that witness (spatial 0.92, object 0.90, goal 0.92, LIBERO-10 0.73). Base pose and table height are witnessed per direction arm, one signed direction of a condition: raised or lowered for table height, and positive or negative base yaw for base pose, where the witness replays the all-positive and all-negative sign patterns and each scored episode joins the arm of its manifest's yaw sign; this gives 40 base-pose arms and 20 table-height arms over their 20 and 10 conditions. The witness passes 14 of 20 control-frequency conditions, 26 of 40 base-pose arms, and 10 of 20 table-height arms. Camera axes and proprioception noise are solvable by construction since the physics is untouched. Gripper-pad survival is a covariate, never a filter, because a replayed demonstration cannot adapt its grasp (the anti-selection principle: a witness gates admission only if the witness can adapt).

\paragraph{Admission audit and the raw-scoring decision.}
We first used the certificates as an admission gate. An audit of the 30 units the witness could not certify (6 control-frequency conditions, 14 base-pose direction arms, 10 table-height arms) classified 29 as solvable by existence proof: some evaluated model reached at least 30 percent of its own nominal on exactly those episodes, and on 17 of them some model exceeded 0.85 absolute. On 28 of 30 the best model outscored the witness itself; on the goal-suite table-height conditions at the lowest levels, even the weakest model scored 0.94 to 1.00 where the witness scored 0.63. The one remaining unit (base pose, goal suite, harshest level) still had 20 successes across four models and is hard, not proven unsolvable. Rejected units did score lower than admitted ones (for OpenVLA-OFT, 0.610 against 0.747 relative to nominal), which is exactly the survivorship asymmetry: dropping them can only raise the mean. Across the four baseline models the gate had raised the retention-based axis-mean score then in use by 0.007 to 0.016, inflated scores for the less robust models more, flipped the \pizero and \pihalf order on the table-height axis, and deleted the most discriminative conditions (spatial control frequency level 4: OpenVLA-OFT 0.02 against \pihalf 0.38; goal level 4: 0.24 against 0.64, in the audit's screen-tier data; 0.02 against 0.32 in the confirm tier). Twenty-five of the 30 rejections came from the goal suite, where the witness sits at a flat 0.58 to 0.68 ceiling at every severity, including level 1 where every model scores 0.91 to 1.00: a tracker limitation, not a task property. The protocol therefore counts all 210 conditions and every episode for every model, and the certificates, witness rates, and blind-episode lists will be released as diagnostics with the results.

\paragraph{Camera visibility audit.}
For all four camera axes, one demonstration per task is played back under every perturbed camera direction with segmentation rendering of the perturbed camera, the wrist camera for the wrist-mount axis (800 records, 3,600 scene-camera and 3,200 wrist-camera directions; frames sampled at stride 10; a goal object counts as visible in a frame if it covers at least 15 pixels). The per-direction worst-object visibility is sharply bimodal: a direction either keeps every goal object visible most of the time or a goal object is essentially never visible. A (task, condition, direction) combination is flagged blind if some goal-relevant object is visible in fewer than 5 percent of sampled frames, and the flags transfer to the scored episodes through the manifest's direction; any threshold between 3 and 10 percent changes the set by at most 6 of 315 combinations. The 315 flagged combinations map to 2,017 of the 40,000 camera-axis episodes (5.0 percent; the four camera axes hold 40,000 confirm episodes per full-coverage model, and the wrist axis contributes no flagged combination, its minimum worst-object visibility staying above 0.1 at every severity), concentrated in camera position levels 3 to 5 and camera orientation level 5, with the object suite's shared pick-and-place basket the most common source of visibility failures; 97 percent of the offending objects never reach the pixel threshold in any frame, so the object is out of frame or occluded from the start rather than hidden by the task. The wrist-camera axis has no blind direction at any severity. The list is published with the benchmark and is not applied to the score. Two facts survive from it. The camera collapses are behavioral: OpenVLA-OFT falls to 0.23 on the goal suite at camera position level 4 with zero blind directions, and in the worst long-horizon camera conditions the baseline's success on provably sighted episodes is at most 13 percent (screen tier). And a policy's success on an episode where the goal object is invisible to the moved camera cannot be read as memorization, because the wrist camera is unperturbed on that axis and not recorded, so wrist-guided placement and memorized geometry predict the same observation; we withdrew an earlier claim to that effect.

\section{The Excluded Axis: Actuation Latency}
\label{sec:supp_latency}
An actuation-latency axis (commands executed 1, 2, 3, 5, or 8 control steps late, 50 to 400 ms) was part of the pilot design. It is the most destructive fault we measured: 400 ms of delay took the OpenVLA-OFT baseline from 97 percent to 1 percent in the pilot screen, and a compensating replay witness solved every latency condition. It was cut from the score because a fix exists entirely in the software around the policy: an executor that waits until the arm has settled and then re-plans, instead of executing a stale action chunk, removes the fault without modifying the policy, and LIBERO's quasi-static scenes permit this intervention. An axis that a systems engineer can zero out without touching the model measures the executor, not the policy. The data will be released as a separate system track, and the level-5 latency condition survives as a parity-gate condition (1 of 100). Destructive is not the same as fair to ask.

\section{Model Setup and Parity}
\label{sec:supp_models}
\cref{tab:models} lists every arm. All models except the two OpenVLA-OFT variants run as policy servers behind a single unchanged client that speaks the openpi wire protocol; the OpenVLA-OFT variants run in-process through the authors' evaluation code with a perturbation hook. Before entering the benchmark, each model ran a nominal preflight of 400 episodes under its authors' settings (100 for the LIBERO-10-only arm; a separate run for the arms added later, the screen tier's nominal conditions for the earlier arms), read against its published average with 0.85 as the guideline. \cref{tab:models} lists the published average beside our measured confirm nominal. The gaps (NORA 87.9 against 83.8, MolmoAct 86.6 against 84.5, LingBot-VA 98.5 against 95.0 on LIBERO-10) have not been isolated to a cause. The documented differences are our image preprocessing (NORA), the serving stack (MolmoAct's published number comes from a vLLM variant), and a single seed against a multi-seed average (Cosmos Policy); none has been ablated. NORA and MolmoAct entered below the guideline, at 0.818 and 0.848 in their 400-episode preflight, after integration checks. The preflight is a parity and integration check: it caught the X-VLA delta-mode collapse described below. Per-model notes follow.

\textbf{OpenVLA-OFT and OFT-Plus.} Released joint four-suite checkpoint; L1 regression head; eight-action chunk executed fully; center crop; state used. OFT-Plus is the LIBERO-Plus authors' fine-tune of the same architecture on their perturbation mix (model card name OpenVLA-OFT+), served identically.
\textbf{\pizero and \pihalf.} Official openpi LIBERO checkpoints and preprocessing (256 px render rotated 180 degrees, resize with padding to 224, 8-dimensional state); ten flow steps; the authors' replan interval of 5 despite horizons of 50 and 10; a seeded server variant so that confirm-tier episodes are reproducible.
\textbf{X-VLA.} Joint LIBERO checkpoint (0.9B) with its bundled remote code; ten denoise steps; served with its native absolute-pose actions (the controller switched to absolute mode after the settle steps), 30-step chunks executed fully, and a constant 90-degree correction between the hand body frame and the grip site. An earlier run that converted absolute actions to deltas with replan 5 collapsed (nominals 0.26, 0.01, 0.57, 0.03, every failure a timeout at half the intended pace) and was discarded as an integration error, not a result.
\textbf{NORA.} The NORA-Long per-suite fine-tunes; greedy decoding of FAST action tokens into five-action chunks; scene camera only, as in the authors' harness. Images take the shared client's 180-degree rotation and bilinear resize to 224 pixels, whereas the authors' OpenVLA-derived helper, reconstructed because the helper files in the checked-out release are empty, uses a JPEG round trip and Lanczos resizing; the effect of this difference is unmeasured. The authors' evaluation also decodes greedily with seed 7, so decoding and seeding do not explain the gap. By suite, published against ours: spatial 92.2 against 89.0, object 95.4 against 91.8, goal 89.4 against 87.8, LIBERO-10 74.6 against 66.4.
\textbf{MolmoAct.} Per-suite checkpoints; the authors' preprocessing (JPEG round trip, 256 px, 0.9 center crop), fixed reasoning template, greedy decoding with up to 512 new tokens, eight-action chunks executed fully; state ignored, as in the authors' harness. Its published average comes from a vLLM variant of the same recipe. Forty-eight screen units that had recorded zero successes because a serving fault dropped the connection were quarantined and rerun; four confirm units that finished through a queue-bypass runner lack video and action logs.
\textbf{Cosmos Policy.} The joint LIBERO checkpoint fine-tuned from Cosmos-Predict2-2B-Video2World; five denoise steps; 16-action chunks executed fully; deterministic inference following the authors' seeding; one seed against the authors' three-seed average.
\textbf{LingBot-VA.} The LIBERO-10 post-trained checkpoint; a stateful video-action protocol with one request per control step and 16-action chunks (four frames of four actions), replayed server-side with episode boundaries detected from seed continuity; images downsampled from the 256 px render to the model's 128 px input.

\begin{table*}[t]
\centering
\caption{\textbf{Evaluated models: checkpoints, interfaces, and parity.} Chunk is the number of actions produced per policy query; replan is the number executed before the next query. Inputs: S scene camera, W wrist camera, P proprioception passed into the network. Published is the model's own reported LIBERO average success in percent where the authors report one (for OpenVLA-OFT, the authors' figure for this combined four-suite checkpoint, 96.8, rather than the 97.1 of their per-suite checkpoints); Ours is the unperturbed success in percent that we measured under our evaluation protocol, with the inference settings and adaptations listed in \cref{sec:supp_models} (confirm tier, averaged over the suites the checkpoint covers). $^\dagger$LIBERO-10 only.}
\label{tab:models}
\scriptsize
\setlength{\tabcolsep}{3pt}
\begin{adjustbox}{max width=\textwidth}\begin{tabular}{@{}l l l l l c c@{}}
\toprule
Model & Checkpoint & Decoding family & Action interface (space; chunk; replan) & Inputs & Published & Ours \\
\midrule
OpenVLA-OFT & \texttt{moojink/openvla-7b-oft-finetuned-libero-spatial-object-goal-10} & regression chunk & delta pose; 8; 8 & S, W, P & 96.8 & 96.6 \\
OFT-Plus & \texttt{Sylvest/openvla-7b-oft-finetuned-libero-plus-mixdata} & regression chunk & delta pose; 8; 8 & S, W, P & -- & 97.0 \\
\pizero & \texttt{openpi} \texttt{pi0\_libero} & flow matching & delta pose; 50; 5 & S, W, P & -- & 91.9 \\
\pihalf & \texttt{openpi} \texttt{pi05\_libero} & flow matching & delta pose; 10; 5 & S, W, P & 96.9 & 96.9 \\
X-VLA & \texttt{2toINF/X-VLA-Libero} & flow matching & absolute pose; 30; 30 & S, W, P & 98.1 & 97.6 \\
NORA & \texttt{declare-lab/nora-long-finetuned-libero-\{spatial,object,goal,10\}} & autoregressive tokens & delta pose; 5; 5 & S & 87.9 & 83.8 \\
MolmoAct & \texttt{allenai/MolmoAct-7B-D-LIBERO-\{Spatial,Object,Goal,Long\}-0812} & autoregressive tokens with trace & delta pose; 8; 8 & S, W & 86.6 & 84.5 \\
Cosmos Policy & \texttt{nvidia/Cosmos-Policy-LIBERO-Predict2-2B} & world-model diffusion & delta pose; 16; 16 & S, W, P & 98.5 & 97.3 \\
LingBot-VA$^\dagger$ & \texttt{robbyant/lingbot-va-posttrain-libero-long} & world-model, video and action & delta pose; 16; stateful, 1 per step & S, W & 98.5 (LIBERO-10) & 95.0 \\
\bottomrule
\end{tabular}\end{adjustbox}
\end{table*}

\section{Sources for the Benchmark Comparison}
\label{sec:supp_sources}
\cref{tab:sources} records, for each benchmark in \cref{tab:benchmarks}, the version we checked and which factors its reported evaluation varies. The comparison is limited to these five benchmarks; it does not claim that no other work has varied a given factor, and it separates a released benchmark's evaluation from a simulator capability and from training-time randomization.
\begin{table*}[t]
\centering
\caption{\textbf{What existing robustness benchmarks perturb.} A check marks a factor that the benchmark's reported evaluation varies (versions and sources in \cref{sec:supp_sources}). The compared benchmarks change the task or the scene; of the rig they vary at most the third-person camera and its optics, the arm's start configuration (LIBERO-Plus), or the table position (Factor World, whose table offsets reach $-5$ to $+2.5$ cm vertically). Factor World also implements initial-arm-pose variation, which its reported study excludes. COLOSSEUM alone varies object mass and friction, physical properties of the objects rather than of the rig. \rig varies only rig properties.}
\label{tab:benchmarks}
\setlength{\tabcolsep}{4pt}
\small
\begin{adjustbox}{max width=\textwidth}\begin{tabular}{@{}l l c c c c c c c c c c@{}}
\toprule
Benchmark & Environment & \rot{task or scene} & \rot{object mass, friction} & \rot{camera pose} & \rot{camera optics} & \rot{wrist camera} & \rot{camera-to-arm calib.} & \rot{actuation gain} & \rot{control rate} & \rot{proprioception} & \rot{base, table, gripper} \\
\midrule
LIBERO-PRO~\cite{zhou2025liberopro} & LIBERO (robosuite) & \checkmark & -- & -- & -- & -- & -- & -- & -- & -- & -- \\
LIBERO-Plus~\cite{fei2026liberoplus} & LIBERO (robosuite) & \checkmark & -- & \checkmark & \checkmark & -- & -- & -- & -- & -- & -- \\
COLOSSEUM~\cite{pumacay2024colosseum} & RLBench (CoppeliaSim) & \checkmark & \checkmark & \checkmark & -- & -- & -- & -- & -- & -- & -- \\
Factor World~\cite{xie2023factorworld} & Meta-World (MuJoCo) & \checkmark & -- & \checkmark & -- & -- & -- & -- & -- & -- & table \\
VLABench~\cite{zhang2024vlabench} & dm\_control (MuJoCo) & \checkmark & -- & -- & -- & -- & -- & -- & -- & -- & -- \\
\midrule
\rig (ours) & LIBERO (robosuite) & fixed & -- & \checkmark & \checkmark & \checkmark & \checkmark & \checkmark & \checkmark & \checkmark & \checkmark \\
\bottomrule
\end{tabular}\end{adjustbox}
\end{table*}

\begin{table*}[t]
\centering
\caption{\textbf{Sources for \cref{tab:benchmarks}.} For each compared benchmark, the version checked, the rig-related factors its reported evaluation varies, factors implemented but excluded from that evaluation, and its other factors. A dash means the checked source documents no such factor; it is not proof that the software cannot express one.}
\label{tab:sources}
\small
\setlength{\tabcolsep}{5pt}
\begin{adjustbox}{max width=\textwidth}\begin{tabular}{@{}l l p{4.2cm} p{3.4cm} p{5.6cm}@{}}
\toprule
Benchmark & Version checked & Rig-related factors evaluated & Implemented, not evaluated & Other factors \\
\midrule
LIBERO-PRO & arXiv 2510.03827v2 & -- & -- & object position, object appearance, instruction, task goal, visual environment (background, lighting, texture) \\
LIBERO-Plus & CVPR 2026 paper & third-person camera viewpoint and field of view; arm start configuration & -- & lighting, background texture, image noise, object layout, language \\
COLOSSEUM & RSS 2024 (arXiv 2402.08191) & pose of the front and two shoulder cameras & -- & object and table color, texture, and size; lighting; background; distractors; object mass and friction \\
Factor World & arXiv 2307.03659 & camera position; table position ($x$, $y$ up to $\pm$7.5 cm; $z$ from $-5$ to $+2.5$ cm) & initial arm pose; object size and shape & lighting, distractors, table and floor texture, object texture, object position \\
VLABench & arXiv 2412.18194 & -- & -- & objects, scenes, language, long-horizon task composition \\
\bottomrule
\end{tabular}\end{adjustbox}
\end{table*}

\section{Compute and Provenance}
\label{sec:supp_compute}
All evaluations ran on nodes with eight NVIDIA H100 GPUs, the cluster's 80 GB variant; every policy server fit on one GPU, so the memory capacity reflected available hardware rather than a requirement. A node typically ran one policy server on one GPU and seven simulation workers on the rest (four suite servers and four workers for the per-suite autoregressive models). The confirm tier evaluates 2,140 units (condition, task) per four-suite model, \episodesPerArm{} episodes each, and \episodesTotal{} episodes over the nine arms; the screen tier is 21,400 episodes per four-suite arm (5,100 for LingBot-VA). We count a node as occupied from its first to its last logged episode, so node-hours include idle tails and reruns, and GPU-hours are eight times node-hours; \texttt{scripts/compute\_hours.py} recomputes every figure below from the event logs. Measured this way, the confirm tier cost 36 node-hours each for OpenVLA-OFT and OFT-Plus, 40 for X-VLA, 63 for \pihalf, 64 for \pizero, 167 for Cosmos Policy, 272 for NORA, 202 for LingBot-VA on its single suite, and about 5,900 for MolmoAct: roughly 6,800 node-hours, or 54,000 GPU-hours, in all. Wall-clock cost is dominated by inference speed. MolmoAct's autoregressive decoding with a reasoning trace completed about 0.4 units per node-hour, its 100 Hz units took a median of 22 hours of worker time each (6.5 to 96 hours), and its confirm tier ran for 11 days over 95 node allocations; across all arms, up to 80 nodes were active at once. Every confirm episode has a video, the commanded action sequence with its seeds and budgets, and an event log, with one exception: four MolmoAct units (200 episodes) finished through a queue-bypass runner that recorded success counts only. A full arm's videos are about 107,000 files and 8 to 11 GB. Results are archived with the frozen manifests and checksums, and the leaderboard is rebuilt from the per-condition files by a single script.

\section{Rig-Augmentation Pilot}
\label{sec:supp_rigaug}
Before the benchmark was frozen, we fine-tuned OpenVLA-OFT on its LIBERO demonstrations replayed under randomly sampled camera rigs, mixing the replayed data at a weight of 0.4 (T1) and 0.3 with a larger long-horizon share (T1b). Both were evaluated on the pilot design of that time (eight axes including latency, ten trials per task) with a retention score, the mean over axes of perturbed success divided by the model's own nominal, so the numbers are not comparable to \cref{tab:leaderboard}. T1 scored 0.76 and T1b about 0.71 against the same-era baseline's 0.79. Camera axes improved on the two short suites where the replayed demonstrations came from (0.95 and above on spatial and object), the latent axes did not move, and the LIBERO-10 nominal fell to 0.69 for T1 (mixture starvation, diagnosed on a checkpoint ladder) and recovered to 0.85 for T1b at the cost of most of the camera gains. Neither arm beat the baseline overall. The lesson matches the OFT-Plus finding in the main text: augmentation improves performance on the trained camera faults, at a cost elsewhere, and does nothing for faults that appear only in the robot's motion.

\end{document}